\documentclass[11pt,english]{article}
\usepackage[T1]{fontenc}
\usepackage[latin9,utf8]{inputenc}
\usepackage{geometry}
\usepackage{color}
\usepackage{babel}
\usepackage{array}
\usepackage{float}
\usepackage{calc}
\usepackage{mathtools}
\usepackage{enumitem}
\usepackage{multirow}
\usepackage{amsmath}
\usepackage{amsthm}
\usepackage{amssymb}
\usepackage{xargs}[2008/03/08]
\usepackage[unicode=true,pdfusetitle,
bookmarks=true,bookmarksnumbered=false,bookmarksopen=false,
breaklinks=false,pdfborder={0 0 1},backref=false,colorlinks=true]
{hyperref}
\hypersetup{
	citecolor=blue, linkcolor=red}

\makeatletter

\providecommand{\tabularnewline}{\\}
\floatstyle{ruled}
\newfloat{algorithm}{tbp}{loa}
\providecommand{\algorithmname}{Algorithm}
\floatname{algorithm}{\protect\algorithmname}

\theoremstyle{plain}
\newtheorem{thm}{\protect\theoremname}
\theoremstyle{plain}
\newtheorem{cor}{\protect\corollaryname}
\theoremstyle{plain}
\newtheorem{fact}{\protect\factname}
\theoremstyle{plain}
\newtheorem{prop}{\protect\propositionname}
\theoremstyle{plain}
\newtheorem{lem}{\protect\lemmaname}
\theoremstyle{plain}
\newtheorem{claim}{\protect\claimname}

\usepackage[mathscr]{euscript}
\usepackage{amsfonts}
\usepackage{dsfont}
\usepackage{times}
\usepackage{amsmath}
\usepackage{bbm}
\pdfmapfile{+txfonts.map}
\usepackage{txfonts}
\usepackage{appendix}
\allowdisplaybreaks

\theoremstyle{definition}
\newtheorem{mdl}{Model}

\usepackage{thm-restate}

\renewcommand{\mathbf}{\boldsymbol}

\usepackage[square,numbers]{natbib}

\makeatother

\providecommand{\factname}{Fact}
\providecommand{\lemmaname}{Lemma}
\providecommand{\propositionname}{Proposition}
\providecommand{\corollaryname}{Corollary}
\providecommand{\theoremname}{Theorem}
\providecommand{\claimname}{Claim}

\begin{document}
\global\long\def\E{\mathbb{E}}%
\global\long\def\P{\mathbb{P}}%
\global\long\def\Var{\operatorname*{Var}}%
\global\long\def\Cov{\operatorname*{Cov}}%
\global\long\def\Tr{\operatorname*{Tr}}%
\global\long\def\diag{\operatorname*{diag}}%
\newcommandx\norm[2][usedefault, addprefix=\global, 1=\#1]{\left\Vert #1\right\Vert {}_{#2}}%
\renewcommandx\norm[2][usedefault, addprefix=\global, 1=\#1]{\Vert#1\|_{#2}}%
\global\long\def\opnorm#1{\norm[#1]{\textup{op}}}%
\global\long\def\rank#1{\operatorname*{rank}(#1)}%
\global\long\def\md#1{\operatorname*{maxdiag}(#1)}%
%\global\long\def\indic{\operatorname*{\mathbbm{1}}}%
\global\long\def\indic{\operatorname*{\mathbb{I}}}%
\global\long\def\diff{\operatorname{d}\!}%
\global\long\def\argmax{\operatorname*{arg\,max}}%
\global\long\def\argmin{\operatorname*{arg\,min}}%
\global\long\def\Bern{\operatorname{Bern}}%
\global\long\def\pos{\operatorname{pos}}%
\global\long\def\round{\operatorname{round}}%

\global\long\def\mtx#1{\bm{#1}}%
\global\long\def\vct#1{\bm{#1}}%
\global\long\def\real{\mathbb{R}}%

\global\long\def\A{\mathbf{A}}%
\global\long\def\Atilde{\widetilde{\A}}%
\global\long\def\B{\mathbf{B}}%
\global\long\def\C{\mathbf{C}}%
\global\long\def\D{\mathbf{D}}%
\global\long\def\F{\mathbf{F}}%
\global\long\def\G{\mathbf{G}}%
\global\long\def\H{\mathbf{H}}%
\global\long\def\I{\mathbf{I}}%
\global\long\def\J{\mathbf{J}}%
\global\long\def\K{\mathbf{K}}%
\global\long\def\L{\mathbf{L}}%
\global\long\def\M{\mathbf{M}}%
\global\long\def\P{\mathbb{P}}%
\global\long\def\O{\mathbf{O}}%
\global\long\def\Q{\mathbf{Q}}%
\global\long\def\R{\mathbf{R}}%
\global\long\def\U{\mathbf{U}}%
\global\long\def\Uperp{\mathbf{U}_{\perp}}%
\global\long\def\V{\mathbf{V}}%
\global\long\def\Vtilde{\widetilde{\mathbf{V}}}%
\global\long\def\Vprime{\mathbf{V}'}%
\global\long\def\W{\mathbf{W}}%
\global\long\def\Wtilde{\widetilde{\W}}%
\global\long\def\X{\mathbf{X}}%
\global\long\def\Y{\mathbf{Y}}%
\global\long\def\Z{\mathbf{Z}}%
\global\long\def\one{\mathbf{1}}%
\global\long\def\calV{\mathcal{V}}%
\global\long\def\calR{\mathcal{R}}%
\global\long\def\calVrow{\mathcal{V}_{\mbox{row}}}%
\global\long\def\calVcol{\mathcal{V}_{\mbox{col}}}%
\global\long\def\x{\mathbf{x}}%
\global\long\def\y{\mathbf{y}}%
\global\long\def\a{\mathbf{a}}%
\global\long\def\b{\mathbf{b}}%
\global\long\def\e{\mathbf{e}}%
\global\long\def\bzero{\mathbf{0}}%

\global\long\def\Yhat{\widehat{\Y}}%
\global\long\def\Ystar{\Y^{*}}%
\global\long\def\Adj{\A}%
\global\long\def\ObsAdj{\B}%
\global\long\def\Noise{\W}%
\global\long\def\HalfAdj{\boldsymbol{\Lambda}}%
\global\long\def\HalfNoise{\boldsymbol{\Psi}}%
\global\long\def\Yhp{\widehat{\Y}_{\perp}}%
\global\long\def\Yround{\widehat{\Y}^{\text{R}}}%
\global\long\def\CensorMat{\Z}%
\global\long\def\bOmega{\boldsymbol{\Omega}}%
\global\long\def\Meanhat{\hat{\boldsymbol{\mu}}}%
\global\long\def\Mean{\boldsymbol{\mu}}%
\global\long\def\mean{\mu}%
\global\long\def\g{\mathbf{g}}%
\global\long\def\h{\mathbf{h}}%
\global\long\def\w{\mathbf{w}}%
\global\long\def\u{\mathbf{u}}%
\global\long\def\v{\mathbf{v}}%
\global\long\def\Fhat{\widehat{\F}}%
\global\long\def\Fstar{\F^{*}}%
\global\long\def\fhat{\widehat{F}}%
\global\long\def\fstar{F^{*}}%

\global\long\def\yhat{\widehat{Y}}%
\global\long\def\ystar{Y^{*}}%
\global\long\def\adj{A}%
\global\long\def\obsadj{B}%
\global\long\def\noise{W}%
\global\long\def\halfadj{\Lambda}%
\global\long\def\halfnoise{\Psi}%
\global\long\def\yround{\widehat{Y}^{\text{R}}}%
\global\long\def\censormat{Z}%

\global\long\def\OneMat{\J}%
\global\long\def\onemat{J}%
\global\long\def\onevec{\mathbf{1}}%
\global\long\def\IdMat{\I}%
\global\long\def\CovMat{\boldsymbol{\Sigma}}%

\global\long\def\num{n}%
\global\long\def\size{\ell}%
\global\long\def\numclust{k}%
\global\long\def\snr{s}%
\global\long\def\inprob{p}%
\global\long\def\outprob{q}%
\global\long\def\flipprob{\epsilon}%
\global\long\def\obsprob{\alpha}%
\global\long\def\apxconst{\rho}%
\global\long\def\error{\gamma}%
\global\long\def\LabelStar{\boldsymbol{\sigma}^{*}}%
\global\long\def\labelstar{\sigma^{*}}%
\global\long\def\LabelHat{\widehat{\boldsymbol{\sigma}}}%
\global\long\def\labelhat{\widehat{\sigma}}%
\global\long\def\z{\mathbf{z}}%
\global\long\def\vecdim{d}%
\global\long\def\misrate{\operatorname*{\texttt{err}}}%
\global\long\def\clustering{\texttt{cluster}}%
\global\long\def\minsep{\Delta}%
\global\long\def\sepratio{\lambda}%
\global\long\def\std{\nu}%
\global\long\def\sgnorm{\tau}%

\global\long\def\PT{\mathcal{P}_{T}}%
\global\long\def\PTperp{\mathcal{P}_{T^{\perp}}}%
\global\long\def\positify{{\cal T}}%
\global\long\def\clustset#1{C_{#1}^{*}}%
\global\long\def\clustest#1{\widehat{C}_{#1}}%
\global\long\def\vertexset{V}%
\global\long\def\perm{\pi}%
\global\long\def\poly{\text{poly}}%

\global\long\def\constsumthreeterms{72}%
\global\long\def\constoperatornormW{182}%
\global\long\def\t{{\displaystyle ^{\top}}}%
\global\long\def\Abar{\bar{A}}%
\global\long\def\bbar{\bar{b}}%
\global\long\def\ybar{\bar{y}}%
\global\long\def\xbar{\bar{x}}%
\global\long\def\xtilde{\tilde{x}}%
\global\long\def\constp{C_{\inprob}}%
\global\long\def\consts{C_{\snr}}%
\global\long\def\conste{C_{e}}%
\global\long\def\constu{c_{u}}%
\global\long\def\constdense{c_{d}}%
\global\long\def\constgamma{C_{g}}%
\global\long\def\constmis{C_{m}}%
\global\long\def\constb{C_{b}}%

\global\long\def\pairset{\mathcal{L}}%

\global\long\def\calM{\mathcal{M}}%

\global\long\def\calG{\mathcal{G}}%

\global\long\def\calS{\mathcal{S}}%

\global\long\def\calA{\mathcal{A}}%
\global\long\def\calX{\mathcal{X}}%
\global\long\def\calB{\mathcal{B}}%

\global\long\def\IP{\textup{\texttt{IP}}}%
\global\long\def\iperror{\gamma_{\IP}}%

\def\cmt{\color{black}}  % for publication

%%%%%%%%%%%%%%%%
% Redefine environment
%\renewenvironment{proof}
%{\proof{Proof. }}
%{\endproof}
%%%%%%%%%%%%%%%%

\title{Hidden Integrality and Semi-random Robustness of SDP Relaxation for
	Sub-Gaussian Mixture Model}

\author{Yingjie Fei$^1$ and Yudong Chen$^2$\\
	\normalsize 
	$^1$ School of Operations Research and Information Engineering, Cornell University \\
	\normalsize 
	$^2$ Department of Computer Sciences, University of Wisconsin-Madison\\
	\normalsize 
	yf275@cornell.edu, yudong.chen@wisc.edu}

\date{}
\maketitle

\begin{abstract}
We consider the problem of estimating the discrete clustering structures
under the Sub-Gaussian Mixture Model. Our main results establish a
\emph{hidden integrality} property of a semidefinite programming (SDP)
relaxation for this problem: while the optimal solution to the SDP
is not integer-valued in general, its estimation error can be upper
bounded by that of an idealized integer program. The error of the
integer program, and hence that of the SDP, are further shown to decay
\emph{exponentially} in the signal-to-noise ratio. In addition, we
show that the SDP relaxation is robust under the semi-random setting
in which an adversary can modify the data generated from the mixture
model. In particular, we generalize the hidden integrality property
to the semi-random model and thereby show that SDP achieves the optimal
error bound in this setting. These results together highlight the
``global-to-local'' mechanism that drives the performance of the
SDP relaxation.

To the best of our knowledge, our result is the first exponentially
decaying error bound for convex relaxations of mixture models. A corollary
of our results shows that in certain regimes the SDP solutions are
in fact integral and exact. More generally, our results establish
sufficient conditions for the SDP to correctly recover the cluster
memberships of $(1-\delta)$ fraction of the points for any $\delta\in(0,1)$.
As a special case, we show that under the $\vecdim$-dimensional Stochastic
Ball Model, SDP achieves non-trivial (sometimes exact) recovery when
the center separation is as small as $\sqrt{1/\vecdim}$, which improves
upon previous exact recovery results that require constant separation.
\end{abstract}

%\maketitle

\section{Introduction\label{sec:intro}}

We consider the Sub-Gaussian Mixture Model (SGMM), in which one is
given $\num$ random points drawn from a mixture of $\numclust$ sub-Gaussian
distributions with different means/centers. SGMM, particularly its
special case the Gaussian Mixture Model (GMM), is widely used in a
broad range of applications including speaker identification, background
modeling and online recommendation. In these applications, one is
typically interested in two types of statistical inference problems
under SGMM:
\begin{itemize}
	\item \textbf{Clustering:} (approximately) identify the cluster membership
	of each point, that is, which of the $\numclust$ mixture components
	generates a given point;
	\item \textbf{Parameter estimation:} estimate the parameters (e.g., means/centers)
	of the $\numclust$ components, or the density of the entire mixture.
\end{itemize}
Standard approaches to these problems, such as k-means clustering,
typically lead to integer programming formulations that are non-convex
and NP-hard to optimize~\cite{aloise2009hard,jain2002facility,mahajan2009planar}.
Consequently, much work has been devoted to developing computationally
tractable algorithms for SGMM; examples include expectation maximization
\cite{dempster1977em}, Lloyd's algorithm \cite{lloyd1982least},
spectral methods \cite{vempala2004spectral}, the method of moments
\cite{pearson1936method}, and many more. Among them, the convex relaxation
methods, including those based on linear programming (LP) and semidefinite
programming (SDP), have emerged as a promising approach for SGMM.
This approach has several attractive properties: (a) it is solvable
in polynomial time, and does not require a good initial solution to
be provided; (b) it has the flexibility to incorporate different quality
metrics and additional constraints; (c) it is not restricted to specific
forms of SGMM (such as Gaussian distribution), and is robust against
model misspecification \cite{peng2005new,peng2007approximating,nellore2015recovery};
(d) it can provide a certificate for optimality \cite{iguchi2017certify}.

Theoretical performance guarantees for convex relaxation methods have
been investigated in a body of both old and recent work. As will be
discussed in greater details in the related work section (Section
\ref{sec:related}), these existing results often come in one of two
forms:
\begin{enumerate}
	\item How well the (rounded) solution of a relaxation optimizes a particular
	objective function (e.g., the k-means or k-medians objective) compared
	to the original integer program, as studied in classical work on \emph{approximation
		factors} \cite{charikar1999constant,kanungo2004local,peng2007approximating,li2016approximating};
	\item When the solution of a relaxation exactly recovers the ground-truth
	clustering, a phenomenon known as \emph{exact recovery} and studied
	in a more recent line of work \cite{nellore2015recovery,awasthi2015relax,mixon2017clustering,iguchi2017certify,li2017kmeans}.
\end{enumerate}
In many applications, optimizing a particular objective function,
and designing approximation algorithms for doing so, are often only
a means to an end, whereas the ultimate goal is to solve the two statistical
inference problems above, both of which involve learning the true
underlying model that generates the observed data. Results on exact
recovery are more directly relevant to this goal. However, such results
often require very stringent conditions on the separation or signal-to-noise
ratio (SNR) of the model. In practice, convex relaxation solutions
are rarely exact, even when the data are generated from the assumed
model. On the other hand, researchers have observed that the solutions,
while not exact or integer-valued, are often a good approximation
to the desired solution that represents the true model~\cite{mixon2017clustering}.
Such a phenomenon is not captured by the results on exact recovery.

In this paper, we aim to strengthen our understanding of the convex
relaxation approach for clustering SGMM. In particular, we study the
regime where solutions of convex relaxations are not exact, and directly
characterize the \emph{estimation errors} of the solutions\textemdash namely,
their distance to desired (integer) solution corresponding to the
true underlying model.

\subsection{Our Contributions}

For a class of SDP relaxations for SGMM, our results reveal a perhaps
surprising property thereof: while the SDP solutions are not integral
in general, their estimation errors can be controlled by that of the
solutions of an idealized integer program (IP), in which one tries
to estimate cluster memberships when an oracle reveals the \emph{true
	centers} of the SGMM. We refer to the latter program as the \emph{Oracle
	Integer Program}. In particular, we show that, in a precise sense
to be formalized later, the estimation errors of the SDP and Oracle
IP satisfy the relationship (Theorem~\ref{thm:ip_sdp})
\begin{equation}
\text{error(SDP)}\lesssim\text{error(IP)}\label{eq:sdp_ip_informal}
\end{equation}
under certain conditions. We refer to this property as \emph{hidden
	integrality} of the SDP relaxation; its proof in fact involves showing
that certain intermediate linear optimization problems are integral.
We then further upper bound the error of the Oracle IP and show that
it decays \emph{exponentially }in terms of the SNR (Theorem~\ref{thm:ip_rate}):
\begin{equation}
\text{error(IP)}\lesssim\exp\left[-\Omega\left(\text{SNR}^{2}\right)\right],\label{eq:ip_rate_informal}
\end{equation}
where the SNR is defined as the ratio of the center separation and
the standard deviation of the mixture components. Combining these
two results immediately leads to explicit bounds on the error of the
SDP solutions (Corollary~\ref{cor:SDP_rate}).

\subsubsection{Robustness under Semi-random Model\label{sec:intro_semi}}

Our results can be generalized to a so-called \emph{semi-random }version
of SGMM. In this setting, an adversary is allowed to modify the data
points generated from SGMM in an arbitrary and potentially adversarial
way (subject to certain monotonicity constraints). This semi-random
setting captures unpredictable deviations from the nominal SGMM\textemdash which
is common in real data\textemdash and it is well recognized to be
much more challenging than the original, purely random model. In fact,
many existing algorithms provably fail in the semi-random setting
\cite{krivelevich2006coloring,awasthi2017clustering}.

We show that SDP relaxation has an inherent robustness property under
the semi-random model. In particular, we establish the following error
bounds:
\begin{equation}
\text{error(SDP)}\lesssim\text{error(IP)}+\varepsilon\qquad\text{and}\qquad\text{error(IP)}\lesssim\exp\left[-\Omega\left(\text{SNR}^{2}\right)\right],\label{eq:semi_rate_informal}
\end{equation}
where $\varepsilon$ denotes the additional error induced by the adversary;
see Theorems~\ref{thm:ip_sdp_semi} and~\ref{thm:ip_rate_semi}
for the precise statement of this result and the expression for $\varepsilon$.
In certain regimes, the error $\varepsilon$ is dominated by $\exp[-\Omega(\text{SNR}^{2})]$,
the error of the Oracle IP, hence the error of the SDP is (order-wise)
unaffected under the semi-random model. In other regimes, the additional
error $\varepsilon$ can be shown to be fundamentally unavoidable.
Note that the adversary only affects the error of the SDP relative
to the Oracle IP, but not the error of the Oracle IP itself.

\subsubsection{Consequences\label{sec:intro_consequence}}

When the SNR is sufficiently large, the above results imply that the
SDP solutions are integral and exact up to numerical errors, hence
recovering existing results on exact recovery as a special case. If
the SNR is low and the SDP solutions are fractional, one may obtain
an explicit clustering from the SDP solutions via a simple, optimization-free
rounding procedure. We show that the error of this explicit clustering
(in terms of the fraction of points misclassified) is also bounded
by the error of the Oracle IP and hence also decays exponentially
in the SNR (Theorem~\ref{thm:cluster_error_rate}). As a consequence,
we obtain sufficient conditions for misclassifying at most $\delta$
fraction of the points for any given $\delta\in[0,1]$. 

Significantly, our results often match and sometimes improve upon
state-of-the-art performance guarantees in settings for which known
results exist, and lead to new guarantees in other less studied settings
of SGMM. For instance, a corollary of our results shows that under
the Stochastic Ball Model, SDP achieves non-trivial (sometimes exact)
cluster recovery even when the center separation $\minsep$ is as
small as $O(\sqrt{1/\vecdim})$, where $\vecdim$ is the dimension
(Section~\ref{sec:ball_model}). For high dimensional settings, this
bound goes beyond existing results that only consider exact recovery
and require $\minsep=\Omega(1)$. We discuss the implications of our
results in details and compare with existing ones after we state the
main theorems.

\subsubsection{The ``Global-to-Local'' Phenomenon \label{sec:global_to_local}}

Our results above are obtained in two steps: (a) relating the SDP
to the Oracle IP, and (b) bounding the Oracle IP errors. This two-step
approach allows us to decouple the two types of mechanisms that determine
the performance of the SDP relaxation: 
\begin{itemize}
	\item On the one hand, step (a) is done by leveraging the structures of
	the \emph{entire} dataset with $\num$ points. In particular, certain
	global spectral properties of the data ensure that the error of the
	SDP is non-trivially bounded (in terms of the Oracle IP error). This
	step is relatively insensitive to the specific structure of the SGMM.
	\item On the other hand, as shall become clear in the sequel, the Oracle
	IP essentially reduces to $\num$ independent clustering problems,
	one for each data point. Knowing the true cluster centers, the Oracle
	IP is optimal in terms of the clustering errors. No other algorithms
	(including SDP relaxations) can achieve a strictly better error, due
	to the inherent randomness of \emph{individual }data points. Step
	(b) above hence captures the local mechanism that determines fine-grained
	error rates as a function of the SNR. 
\end{itemize}
Our analysis establishes the hidden integrality property as the bridge
between these two mechanisms. Our result hence highlights the power
of the SDP approach: it is able to capture the global and local structures
of the data simultaneously, without requiring a good initial solution
or sophisticated pre-processing/post-processing steps.

In the context of the semi-random model, the error bound~(\ref{eq:semi_rate_informal})
shows that the effect of the adversary is restricted to the global
regime in step (a) and does not play a role in the local step (b).
We emphasize that clustering under semi-random models is a highly
challenging problem by its own, and an entire line of work is devoted
to this problem \cite{coja2004coloringSemirandom,feige2001semirandom,krivelevich2006coloring,awasthi2017clustering,makarychev2012approximation}.
Our two-step, hidden integrality-based approach allows for a streamlined
analysis of this setting, as it isolates precisely how the adversary
can impact the clustering error of the SDP.

\subsection{Paper Organization}

The remainder of the paper is organized as follows. In Section~\ref{sec:related},
we discuss related work on SGMM and its special cases. In Section
\ref{sec:setup}, we describe the problem setup for SGMM and the SDP
relaxation approach. In Section \ref{sec:main}, we present our main
theoretical results, discuss their consequences and compare with existing
results. We prove our main theorems in Sections~\ref{sec:proof_ip_sdp}
and~\ref{sec:proof_ip_rate}. The paper is concluded with a discussion
of future directions in Section~\ref{sec:conclusion}. The proofs
of some technical results are deferred to the Appendix.

\section{Related Work\label{sec:related}}

The study of SGMM has a long history and is still an active area of
research. Here we review the most relevant work on algorithms for
SGMM with theoretical performance guarantees. 

The work~\cite{dasgupta1999learning} is among the first to obtain
performance guarantees for GMM. Subsequent work has developed a variety
of algorithms including spectral methods, which achieved improved
guarantees. These results establish sufficient conditions, in terms
of the separation between the cluster centers (or equivalently the
SNR), for achieving (near)-exact recovery of the cluster memberships.
One of the best results is obtained in~\cite{vempala2004spectral}
and requires $\text{SNR}\gtrsim\left(\numclust\ln\num\right)^{1/4}$,
which is later  extended in a long line of work including \cite{achlioptas2005spectral,kumar2010clustering,awasthi2012improved}.
We compare these results with ours in Section~\ref{sec:main}.

Expectation-Maximization (EM) and the closely related Lloyd's algorithm
are two of the most popular methods for GMM. Despite their empirical
effectiveness, non-asymptotic statistical guarantees are established
only recently; see the work in \cite{balakrishnan2017statistical,klusowski2016statistical,yan2017em,lu2016lloyd}.
All these results assume that one has access to a sufficiently good
initial solution, typically obtained by spectral methods. Although
some recent work shows that EM converges from random initialization
under certain restricted settings of GMM \cite{daskalakis2016ten,xu2016em},
for general settings EM is known to suffer from the existence of bad
local minima \cite{jin2016local}. Robustness of the Lloyd's algorithm
under a semi-random GMM is studied in the work \cite{awasthi2017clustering}.

Most relevant to us is work on convex relaxation methods for GMM and
k-means/median problems. A class of SDP relaxations, often called
the \emph{Peng-Wei relaxations}, are developed in the seminal work
in \cite{peng2005new,peng2007approximating}. Thanks to convexity,
these methods do not suffer from the issues of bad local minima faced
by EM/Lloyd's, though it is far from trivial how to round their (typically
fractional) solutions into a valid clustering solution with provable
quality guarantees. In this direction, the work in \cite{awasthi2015relax,iguchi2017certify,li2017kmeans}
establish sufficient conditions for LP/SDP relaxations to achieve
exact recovery, and the work in \cite{mixon2017clustering} proves
approximate recovery error bounds for SDP. These results are directly
comparable to ours. We discuss them in more details in Section \ref{sec:main}
after presenting our main theorems.

In the last decade, the related Stochastic Block Model (SBM) has also
witnessed significant progress on understanding convex relaxation
methods; see \cite{abbe2016recent,li2018convex} for a comprehensive
survey. In particular, much work has been done on exact recovery guarantees
for SDP relaxations of SBM \cite{krivelevich2006coloring,ames2014convex,chen2012sparseclustering,amini2018semidefinite}.
A more recent line of work establishes approximate recovery guarantees
of the SDPs \cite{Guedon2015,MontanariSen16} in the low-SNR regime.
Particularly relevant to us is the work in \cite{fei2017exponential,fei2019SBM,fei2019achieving},
who also establish exponentially decaying error bounds. Interestingly,
these results also highlight a so-called ``local-to-global amplification''
phenomenon \cite{abbe2016recent,abbe2015general,abbe2016exact}, which
is related to what we discussed in Section~\ref{sec:global_to_local}.
Despite the apparent similarity, however, our results on SGMM require
very different analytical techniques, due to the fundamental difference
between the geometric and probabilistic structures of SGMM and SBM.
Moreover, our results establish a general hidden integrality property
of SDP relaxations, which we believe holds more broadly beyond specific
models like SBM and SGMM.

\section{Models and Algorithms\label{sec:setup}}

In this section, we formally set up the clustering problem under SGMM
and describe our SDP relaxation approach.

\subsection{Notations}

Vectors and matrices are denoted by bold letters such as $\mathbf{\u}$
and $\M$. For a vector $\u$, we denote by $u_{i}$ its $i$-th entry.
For a matrix $\M$, we use $\Tr(\M)$ to denote its trace, $M_{ij}$
its $(i,j)$-th entry, $\mbox{diag}\left(\M\right)$ the vector of
its diagonal entries, $\norm[\M]1\coloneqq\sum_{i,j}M_{ij}$ its entry-wise
$\ell_{1}$ norm, $\opnorm{\M}$ its spectral norm (maximum singular
value), $\M_{i\bullet}$ its $i$-th row and $\M_{\bullet j}$ its
$j$-th column. We write $\M\succeq0$ if $\M$ is symmetric positive
semidefinite (psd). The trace inner product between two matrices $\M$
and $\Q$ of the same dimension is denoted by $\left\langle \M,\Q\right\rangle \coloneqq\Tr(\M^{\top}\Q)$.
For a number $a$, $\M\geq a$ means that $M_{ij}\geq a$ for all
entries $(i,j)$. We denote by $\onevec_{m}$ the all-one column vector
of dimension $m$. For a positive integer $i$, let $[i]\coloneqq\{1,2,\ldots,i\}$.
For two non-negative sequences $\{a_{i}\}$ and $\{b_{i}\}$, we
write $a_{i}\lesssim b_{i}$ if there exists a universal constant
$C>0$ such that $a_{i}\le Cb_{i}$ for all $i$, and write $a_{i}\asymp b_{i}$
if $a_{i}\lesssim b_{i}$ and $b_{i}\lesssim a_{i}$.

We recall that the sub-Gaussian norm of a random variable $X$ is
defined as 
\[
\norm[X]{\psi_{2}}\coloneqq\inf\left\{ t>0:\E\exp\big(X^{2}/t^{2}\big)\le2\right\} ,
\]
and $X$ is called sub-Gaussian if $\norm[X]{\psi_{2}}<\infty$. Note
that Gaussian and bounded random variables are sub-Gaussian. Denote
by $\mathbb{S}^{m-1}$ the $m$-dimensional unit $\ell_{2}$ sphere.
A random vector $\x\in\real^{m}$ is sub-Gaussian if for all fixed
vectors $\u\in\real^{m}$, the one dimensional marginal $\left\langle \x,\u\right\rangle $
is a sub-Gaussian random variable. The sub-Gaussian norm of $\x$
is defined as $\norm[\x]{\psi_{2}}\coloneqq\sup_{\u\in\mathbb{S}^{m-1}}\norm[\left\langle \x,\u\right\rangle ]{\psi_{2}}$.
With this notation, a random vector $\x\sim\mathcal{N}(\bzero,\CovMat)$
from the multivariate Gaussian distribution is sub-Gaussian with $\norm[\x]{\psi_{2}}=\opnorm{\CovMat}$. 

\subsection{Sub-Gaussian Mixture Model \label{sec:setup_model}}

We focus on the Sub-Gaussian Mixture Model (SGMM) with balanced clusters.

\begin{mdl}[Sub-Gaussian Mixture Model]\label{mdl:SGMM} Let $\Mean_{1},\ldots,\Mean_{\numclust}\in\real^{\vecdim}$
	be $\numclust$ unknown cluster centers. We observe $\num$ random
	points in $\real^{\vecdim}$ of the form 
	\[
	\h_{i}\coloneqq\Mean_{\labelstar(i)}+\g_{i},\qquad i\in\left[\num\right]
	\]
	where $\labelstar(i)\in[\numclust]$ is the unknown cluster label
	of the $i$-th point, and $\left\{ \g_{i}\right\} $ are zero-mean
	independent sub-Gaussian random vectors with sub-Gaussian norms $\norm[\g_{i}]{\psi_{2}}\le\sgnorm$.\footnote{More explicitly, the sub-Gaussian assumption is equivalent to $\E\exp\big(\langle\g_{i},\v\rangle\big)\le\exp\big(\sgnorm^{2}\norm[\v]2^{2}/2\big),\forall\v\in\real^{\vecdim}.$}
	We assume that the ground-truth clusters have equal sizes, that is,
	$\left|\left\{ i\in\left[\num\right]:\labelstar(i)=a\right\} \right|=\frac{\num}{\numclust}$
	for each $a\in\left[\numclust\right]$.
	
\end{mdl}

Let $\LabelStar\in[\numclust]^{\num}$ be the vector of the true cluster
labels; that is, its $i$-th coordinate is $\labelstar_{i}\equiv\labelstar(i)$
(we use these two notations interchangeably in the paper.) The task
is to estimate the underlying clustering $\LabelStar$ given the observed
data $\{\h_{i}:i\in[\num]\}$.

Note that in Model~\ref{mdl:SGMM} we do not require $\left\{ \g_{i}\right\} $
to be isotropic or identically distributed. This model includes several
important mixture models as special cases:
\begin{itemize}
	\item Spherical GMM, where $\{\g_{i}\}$ are Gaussian with the covariance
	matrix $\sgnorm^{2}\I$. 
	\item More general GMM where the $k$ clusters have non-identical and non-diagonal
	covariance matrices $\left\{ \CovMat_{a}\right\} _{a\in\left[\numclust\right]}$.
	\item The Stochastic Ball Model \cite{nellore2015recovery}, where $\left\{ \g_{i}\right\} $
	are bounded and rotationally invariant random variables; we discuss
	this model in details in Section~\ref{sec:ball_model} 
\end{itemize}
Throughout the paper we assume $\num\ge4$ and $\numclust\ge2$ to
avoid degeneracy. Denote by $\minsep_{ab}\coloneqq\norm[\Mean_{a}-\Mean_{b}]2$
the separation of the centers of clusters $a$ and $b$, and $\minsep\coloneqq\min_{a\ne b\in\left[\numclust\right]}\norm[\Mean_{a}-\Mean_{b}]2$
the minimum separation of the centers. Playing a crucial role in our
results is the quantity 
\begin{equation}
\snr\coloneqq\frac{\minsep}{\sgnorm},\label{eq:snr}
\end{equation}
which is a measure of the signal-to-noise ratio of the SGMM. 

\subsection{Semidefinite Programming Relaxation\label{sec:setup_sdp}}

We now describe our SDP relaxation for clustering SGMM. To begin,
note that each candidate clustering of $\num$ points into $\numclust$
clusters can be represented by an \emph{assignment matrix $\F\in\{0,1\}^{\num\times\numclust}$
}with
\[
F_{ia}=\begin{cases}
1 & \text{if point \ensuremath{i} is assigned to cluster \ensuremath{a}},\\
0 & \text{otherwise.}
\end{cases}
\]
Let $\mathcal{F}\coloneqq\left\{ \F\in\{0,1\}^{\num\times\numclust}:\F\one_{\numclust}=\one_{\num}\right\} $
be the set of all possible assignment matrices. To cluster the points
$\{\h_{i}\}$, a natural approach is to find an assignment $\F$ that
minimizes the total within-cluster pairwise distance. Arranging the
pairwise squared distance as a matrix $\Adj\in\real^{\num\times\num}$
with
\[
\adj_{ij}=\norm[\h_{i}-\h_{j}]2^{2},\qquad\text{for each }(i,j)\in[\num]\times[\num],
\]
we can express the above objective as
\[
\sum_{i,j}\adj_{ij}\indic\{\text{points \ensuremath{i} and \ensuremath{j} are assigned to the same cluster}\}=\sum_{i,j}\adj_{ij}(\F\F^{\top})_{ij}=\left\langle \F\F^{\top},\Adj\right\rangle .
\]
Therefore, the approach described above can be formulated as the integer
program (\ref{eq:kmeans1}) below:
\begin{center}
	\begin{tabular}{ccc}
		\begin{minipage}[t]{0.3\textwidth}%
			\begin{equation}
			\begin{aligned}\min_{\F}\; & \left\langle \F\F^{\top},\Adj\right\rangle \\
			\text{s.t.}\; & \F\in\mathcal{F},\\
			& \one_{\num}^{\top}\F=\frac{\num}{\numclust}\one_{\numclust}^{\top}.
			\end{aligned}
			\label{eq:kmeans1}
			\end{equation}
		\end{minipage} & \hspace{0.1\textwidth} & %
		\begin{minipage}[t]{0.5\textwidth}%
			\begin{equation}
			\begin{aligned}\min_{\mathbf{Y}}\; & \left\langle \Y,\Adj\right\rangle \\
			\text{s.t.}\; & \Y\one_{\num}=\frac{\num}{\numclust}\one_{\num},\\
			& \mbox{diag}\left(\Y\right)=\one_{\num},\\
			& \Y\succeq0,\\
			& \Y\in\left\{ 0,1\right\} ^{\num\times\num},\rank{\Y}=\numclust.
			\end{aligned}
			\label{eq:kmeans2}
			\end{equation}
		\end{minipage}\tabularnewline
	\end{tabular}
	\par\end{center}

In program~(\ref{eq:kmeans1}) the additional constraint $\one_{\num}^{\top}\F=\frac{\num}{\numclust}\one_{\numclust}^{\top}$
enforces that all $\numclust$ clusters have the same size $\frac{\num}{\numclust}$,
as we are working with SGMM whose true clusters are balanced. Under
this balanced model, it is not hard to see that the program~(\ref{eq:kmeans1})
is equivalent to the classical k-means formulation. With a change
of variable $\Y=\F\F^{\top}$, we may lift the program~(\ref{eq:kmeans1})
to the space of $\num\times\num$ matrices and obtain the equivalent
formulation~(\ref{eq:kmeans2}).

Both programs~(\ref{eq:kmeans1}) and~(\ref{eq:kmeans2}) involve
non-convex combinatorial constraints and are computationally hard
to solve. To obtain a tractable formulation, we drop the non-convex
rank constraint in~(\ref{eq:kmeans2}) and replace the integer constraint
with the box constraint $0\le\Y\le1$ (the constraint $\Y\le1$ is
in fact redundant). Doing so leads to the following SDP relaxation:
\begin{equation}
\begin{aligned}\Yhat\in\argmin_{\mathbf{Y}\in\mathbb{R}^{\num\times\num}}\; & \left\langle \Y,\Adj\right\rangle \\
\mbox{s.t.}\; & \Y\one_{\num}=\frac{\num}{\numclust}\one_{\num},\\
& \mbox{diag}\left(\Y\right)=\one_{\num},\\
& \Y\succeq0,\\
& \Y\ge0.
\end{aligned}
\label{eq:SDP1}
\end{equation}

Let $\F^{*}\in\mathcal{F}$ be the assignment matrix associated with
the true underlying clustering of the SGMM; that is, $F_{ja}^{*}=\indic\left\{ \labelstar(j)=a\right\} $
for each $j\in[\num]$ and $a\in[\numclust]$. The performance of
the SDP is measured against the true \emph{cluster matrix} $\Ystar:=\Fstar(\Fstar)^{\top}\in\{0,1\}^{\num\times\num}$,
which has the explicit form
\[
\ystar_{ij}=\begin{cases}
\ensuremath{1} & \text{if \ensuremath{\labelstar(i)=\labelstar(j)}, i.e., points \ensuremath{i} and \ensuremath{j} are in the same cluster},\\
\ensuremath{0} & \text{if \ensuremath{\labelstar(i)\ne\labelstar(j)}, i.e., points \ensuremath{i} and \ensuremath{j} are in different clusters, }
\end{cases}
\]
with the convention that $\ystar_{ii}=1,\forall i\in[\num]$. The
matrix $\Ystar$ hence encodes the ground-truth clustering labels
$\LabelStar$. If we order the rows and columns of $\Ystar$ according
to $\LabelStar$, then $\Ystar$ takes the block-diagonal form
\begin{equation}
\Ystar=\begin{pmatrix}\OneMat_{\num/\numclust}\\
& \ddots\\
&  & \OneMat_{\num/\numclust}
\end{pmatrix},\label{eq:block_diagonal}
\end{equation}
where $\OneMat_{\num/\numclust}$ denotes the $\frac{\num}{\numclust}$-by-$\frac{\num}{\numclust}$
all-one matrix. From Equation~(\ref{eq:block_diagonal}) it is clear
that $\Ystar$ is feasible to the SDP~(\ref{eq:SDP1}). We view an
optimal solution $\Yhat$ to~(\ref{eq:SDP1}) as an estimate of the
true clustering $\Ystar$. Our goal is to characterize the estimation
error $\norm[\Yhat-\Ystar]1$ in terms of the number of points $\num$,
the number of clusters $\numclust$, the data dimension $\vecdim$
and the SNR $\snr$ defined in~(\ref{eq:snr}). Note that here we
measure the error of $\Yhat$ in the $\ell_{1}$ metric. As we shall
see later, this metric is directly related to the clustering error,
i.e., the fraction of misclassified points.

We remark that the SDP~(\ref{eq:SDP1}) is somewhat different from
the more classical SDP relaxation of k-means proposed in \cite{peng2007approximating},
as the latter involves normalizing the variable by the cluster sizes.
The SDP (\ref{eq:SDP1}) has been studied under SGMM in~\cite{li2017kmeans},
and is closely related to the SDP formulation considered in \cite{amini2018semidefinite}
for the Stochastic Block Model.

\subsection{Explicit Clustering \label{sec:setup_explict_clustering}}

In general the SDP solution $\Yhat$ is not in the block-diagonal
form of (\ref{eq:block_diagonal}) and hence does not directly correspond
to an explicit clustering. Nevertheless, we may easily extract cluster
memberships from $\Yhat$ using a simple greedy procedure that runs
in time linear in the size of $\Yhat$.  

The procedure consists of two steps. In the first step, as given in
Algorithm~\ref{alg:apx_clustering}, we treat the rows of $\Yhat$
as points in $\real^{\num}$, and consider the $\ell_{1}$ balls centered
at each row with a certain radius. The ball that contains the most
rows is identified, and the indices of the rows in this ball are output
as a set. The process continues iteratively with the remaining rows
of $\Yhat$. This procedure outputs a collection of index sets whose
sizes are no larger than $\frac{\num}{\numclust}$ but may not equal
to each other. 
\begin{algorithm}[H]
	\caption{Initial Grouping\label{alg:apx_clustering}}
	
	Input: data matrix $\Yhat\in\real^{\num\times\num}$, number of points
	$\num$, target number of clusters $\numclust$.
	\begin{enumerate}
		\item Initialize $B_{0}\leftarrow\emptyset$, $t\leftarrow0$, $\vertexset\leftarrow\left[\num\right]$.
		\item While $\vertexset\backslash\bigcup_{i=0}^{t}B_{i}\ne\emptyset$:
		\begin{enumerate}
			\item Set $t\leftarrow t+1$ and $\vertexset_{t}\leftarrow\vertexset\backslash\bigcup_{i=0}^{t-1}B_{i}$.
			\item For each $u\in\vertexset_{t}$, set $B(u)\leftarrow\left\{ w\in V_{t}:\norm[\Yhat_{u\bullet}-\Yhat_{w\bullet}]1\leq\frac{\num}{4\numclust}\right\} $.
			\item Set $B_{t}\leftarrow\argmax_{B(u):u\in\vertexset_{t}}\left|B(u)\right|$.
			\item If $\left|B_{t}\right|>\frac{\num}{\numclust}$, then remove arbitrary
			elements in $B_{t}$ so that $\left|B_{t}\right|=\frac{\num}{\numclust}$
		\end{enumerate}
	\end{enumerate}
	Output: sets $\left\{ B_{t}\right\} _{t\ge1}$.
\end{algorithm}
In the second step, we convert the sets output above into $\numclust$
equal-size clusters. This is done by identifying the $\numclust$
largest sets among them, and distributing points in the remaining
sets across the chosen $\numclust$ sets so that each of these sets
contains exactly $\frac{\num}{\numclust}$ points. Combining these
two steps gives our final algorithm, $\clustering$, for extracting
an explicit clustering from the SDP solution~$\Yhat$. This procedure
is given as Algorithm~\ref{alg:clustering}. 
\begin{algorithm}[H]
	\caption{$\protect\clustering$ \label{alg:clustering}}
	
	Input: data matrix $\Yhat\in\real^{\num\times\num}$, number of points
	$\num$, target number of clusters $\numclust$.
	\begin{enumerate}
		\item Run Algorithm \ref{alg:apx_clustering} with $\Yhat,\num$ and $\numclust$
		as input and obtain the sets $\left\{ B_{t}\right\} _{t\ge1}$.
		\item Choose the $\numclust$ largest sets among $\left\{ B_{t}\right\} _{t\ge1}$;
		call the chosen sets $\left\{ U_{t}\right\} _{t\in\left[\numclust\right]}$.
		\item Arbitrarily distribute elements of $\left\{ B_{t}\right\} _{t\ge1}\backslash\left\{ U_{t}\right\} _{t\in\left[\numclust\right]}$
		among $\left\{ U_{t}\right\} _{t\in\left[\numclust\right]}$ so that
		each $U_{t}$ has exactly $\frac{\num}{\numclust}$ elements.
		\item For each $i\in\left[\num\right]$: set $\labelhat_{i}\leftarrow t$,
		where $t$ is the unique index in $[\numclust]$ such that $U_{t}\ni i$.
	\end{enumerate}
	Output: clustering assignment $\LabelHat\in[\numclust]^{\num}$.
\end{algorithm}
The output of the above procedure
\[
\LabelHat\coloneqq\clustering(\Yhat,\num,\numclust)
\]
is a vector in $[\numclust]^{\num}$ such that point $i$ is assigned
to the $\labelhat_{i}$-th cluster. We are interested in controlling
the clustering error of $\LabelHat$ relative to the ground-truth
clustering $\LabelStar$. Let $\calS_{\numclust}$ denote the symmetric
group consisting of all permutations of $[\numclust]$. The clustering
error is defined by
\begin{equation}
\misrate(\LabelHat,\LabelStar)\coloneqq\min_{\perm\in\calS_{\numclust}}\frac{1}{n}\left|\left\{ i\in[\num]:\labelhat_{i}\neq\pi(\labelstar_{i})\right\} \right|,\label{eq:mis_rate_def}
\end{equation}
which is the proportion of points that are misclassified, modulo permutation
of the cluster labels. 

Variants of the above $\clustering$ procedure have been considered
before in \cite{makarychev2016learning,mixon2017clustering}. Our
results in the next section establish that the clustering error $\misrate(\LabelHat,\LabelStar)$
is always upper bounded by the $\ell_{1}$ error $\norm[\Yhat-\Ystar]1$
of the SDP solution. 

\section{Main Results \label{sec:main}}

In this section, we establish the connection between the estimation
error of the SDP relaxation~(\ref{eq:SDP1}) and that of what we
call the Oracle Integer Program. Using this connection, we derive
explicit bounds on the error of the SDP, and explore the implications
of these results under different settings.

\subsection{Oracle Integer Program\label{sec:oracleIP}}

Consider an idealized setting where an oracle reveals the true cluster
centers $\left\{ \Mean_{a}\right\} _{a\in\left[\numclust\right]}$.
Moreover, we are given the data points $\big\{\bar{\h}_{i}\big\}_{i\in[\num]}$,
where $\bar{\h}_{i}\coloneqq\Mean_{\labelstar(i)}+(2\kappa)^{-1}\g_{i}$
with $\kappa:=1/8$ and $\left\{ \g_{i}\right\} $ are the same realizations
of the random variables in the original SGMM. In other words, $\big\{\bar{\h}_{i}\big\}$
are the same as the original data points $\big\{\h_{i}\big\}$, except
that the sub-Gaussian norm of the noise $\left\{ \g_{i}\right\} $
is scaled by $(2\kappa)^{-1}=4$. 

To cluster the points $\big\{\bar{\h}_{i}\big\}$ in this setting,
a natural approach is to simply assign each point to the closest cluster
center, so that the total distance of the points to their assigned
centers is minimized. Representing each candidate clustering assignment
using an assignment matrix $\F\in\mathcal{F}$ as before, we see that
the quantity 
\begin{equation}
\eta(\F)\coloneqq\sum_{j\in[\num]}\sum_{a\in[\numclust]}\norm[\bar{\h}_{j}-\Mean_{a}]2^{2}F_{ja}\label{eq:eta_def}
\end{equation}
is exactly the sum of the distances of each point to its assigned
center. The idealized clustering procedure above thus amounts to solving
the following \emph{Oracle Integer Program (IP)}:
\begin{equation}
\begin{aligned}\min_{\F}\; & \eta(\F),\qquad\text{s.t.}\;\F\in\mathcal{F}.\end{aligned}
\label{eq:oracleIP}
\end{equation}
Note that this program is separable across the rows of $\F$ and can
be reduced to $\num$ independent optimization problems, one for each
data point $\bar{\h}_{i}$.

A priori, there is no obvious connection between the estimation error
of a solution to the Oracle IP and that of a solution to the SDP.
In particular, the SDP is oblivious to the true centers and in general
has fractional optimal solutions. Nevertheless, we are able to establish
a formal relationship between these two programs, as is done below.

\subsection{Errors of SDP Relaxation and Oracle IP}

We begin by making the following observations. Recall that $\F^{*}\in\mathcal{F}$
is the true assignment matrix as defined in Section~\ref{sec:setup_sdp}.
For each assignment matrix $\F\in\mathcal{F}$ , it is easy to see
that the quantity $\frac{1}{2}\norm[\F-\F^{*}]1$ equals the number
of points that are assigned differently in $\F$ and $\F^{*}$. Therefore,
this quantity measures the clustering error of $\F$. On the other
hand, for a solution $\F\in\mathcal{F}$ to potentially be an optimal
solution of the Oracle IP~(\ref{eq:oracleIP}), it must satisfy $\eta(\F)\le\eta(\F^{*})$
since $\F^{*}$ is feasible to~(\ref{eq:oracleIP}). Consequently,
the quantity 
\begin{equation}
\max\left\{ \frac{1}{2}\norm[\F-\F^{*}]1:\F\in\mathcal{F},\eta(\F)\le\eta(\F^{*})\right\} \label{eq:IPerror}
\end{equation}
represents the worst-case error of a potentially optimal solution
to the Oracle IP.

The quantity in~(\ref{eq:IPerror}) in fact gives an \emph{upper
	bound} of the error of the optimal solution $\Yhat$ to the SDP relaxation,
as is shown in the theorem below.
\begin{thm}[IP bounds SDP]
	\emph{\label{thm:ip_sdp} }Under Model~\ref{mdl:SGMM}, there exist
	some universal constants $\consts>0,C\ge1$ for which the following
	holds. If the SNR satisfies 
	\begin{equation}
	\snr^{2}\geq\consts\left(\sqrt{\frac{\numclust\vecdim}{\num}\log\num}+\frac{\numclust\vecdim}{\num}+\numclust\right),\label{eq:snr_cond}
	\end{equation}
	then we have
	\[
	\frac{\norm[\Yhat-\Ystar]1}{\norm[\Ystar]1}\le2\cdot\max\left\{ \frac{\norm[\F-\F^{*}]1}{\norm[\F^{*}]1}:\eta(\F)\le\eta(\F^{*}),\F\in\mathcal{F}\right\} 
	\]
	with probability at least $1-\num^{-C}-2e^{-\num}$.
\end{thm}
We prove this theorem in Section~\ref{sec:proof_ip_sdp}. The proof
has two main steps: $(i)$ showing that with high probability the
SDP error is upper bounded by the objective value of a linear program
(LP), and $(ii)$ showing that the LP admits an \emph{integral} optimal
solution and relating this solution to the Oracle IP error in~(\ref{eq:IPerror}).
We note that the key step $(ii)$, which establishes a hidden integrality
property, is completely deterministic. The SNR condition~(\ref{eq:snr_cond})
is required only in the probabilistic step~$(i)$. As we elaborate
in Sections~\ref{sec:consequences}\textendash \ref{sec:ball_model},
our SNR condition holds even in the regime where exact recovery is
impossible, and is often milder than existing results on convex relaxations.
\\

To obtain an explicit bound on the SDP error, it suffices to upper
bound the error~(\ref{eq:IPerror}) of the Oracle IP. This turns
out to be a relatively easy task compared to directly controlling
the error of the SDP: the Oracle IP has only finitely many feasible
solutions, allowing one to use a union-bound-like argument. Our analysis
establishes that the error of the Oracle IP decays \emph{exponentially}
in the SNR, as summarized in the theorem below.
\begin{thm}[Exponential rate of IP]
	\label{thm:ip_rate} Under Model~\ref{mdl:SGMM}, there exist universal
	constants $\consts,\constgamma,\conste>0$ for which the following
	holds. If $\snr^{2}\ge\consts\numclust$, then we have 
	\[
	\max\left\{ \frac{\norm[\F-\F^{*}]1}{\norm[\F^{*}]1}:\eta(\F)\le\eta(\F^{*}),\F\in\mathcal{F}\right\} \le\constgamma\exp\left[-\frac{\snr^{2}}{\conste}\right]
	\]
	with probability at least $1-\frac{3}{2}\num^{-1}$.
\end{thm}
The proof is given in Section \ref{sec:proof_ip_rate}. Combining
Theorems \ref{thm:ip_sdp} and \ref{thm:ip_rate}, we immediately
deduce that the SDP (\ref{eq:SDP1}) also achieves an exponentially
decaying error rate.
\begin{cor}[Exponential rate of SDP]
	\label{cor:SDP_rate} Under Model~\ref{mdl:SGMM} and the SNR condition~(\ref{eq:snr_cond}),
	there exist universal constants $\constmis,\conste>0$ such that 
	\[
	\frac{\norm[\Yhat-\Ystar]1}{\norm[\Ystar]1}\leq\constmis\exp\left[-\frac{\snr^{2}}{\conste}\right]
	\]
	with probability at least $1-2\num^{-1}$.
\end{cor}
Our last theorem concerns the explicit clustering $\LabelHat$ extracted
from $\Yhat$ using the $\clustering$ procedure described in Section
\ref{sec:setup_explict_clustering}. In particular, we show that the
clustering error of $\LabelHat$ is upper bounded by the $\ell_{1}$
error of $\Yhat$; consequently, $\LabelHat$ inherits the exponential
error bound of $\Yhat$.
\begin{thm}[Clustering error]
	\emph{}\label{thm:cluster_error_rate} The clustering $\LabelHat$
	extracted from $\Yhat$ satisfies the deterministic inequality
	\[
	\misrate(\LabelHat,\LabelStar)\lesssim\frac{\norm[\Yhat-\Ystar]1}{\norm[\Ystar]1}.
	\]
	Consequently, under Model~\ref{mdl:SGMM} and the SNR condition~(\ref{eq:snr_cond}),
	there exist universal constants $\constmis,\conste>0$ such that 
	\[
	\misrate\left(\LabelHat,\LabelStar\right)\leq\constmis\exp\left[-\frac{\snr^{2}}{\conste}\right]
	\]
	with probability at least $1-2\num^{-1}$.
\end{thm}
The proof is given in Appendix~\ref{sec:proof_cluster_error_rate}.
Note that the above clustering error bound is information-theoretically
optimal (up to a constant in the exponent) in view of the minimax
results in~\cite{lu2016lloyd}.

{\cmt 
We would like to mention that the SNR condition \eqref{eq:snr_cond} for Theorems \ref{thm:ip_sdp}--\ref{thm:cluster_error_rate} and Corollary~\ref{cor:SDP_rate} can be improved to 
\[
	\snr^{2}\geq\consts\left(\sqrt{\frac{\numclust\vecdim}{\num}\log\num}+\numclust\sqrt{\frac{\vecdim}{\num}}+\numclust\right),
\]
by adopting the proof strategies in the conference version of this paper \citep{fei2018hidden}. We adopt the current proof as it allows us to streamline the analysis for both SGMM (Model \ref{mdl:SGMM}) and the semi-random SGMM (Model \ref{mdl:SGMM_semi} to be introduced below).
}

\subsubsection{Consequences\label{sec:consequences}}

We explore the consequences of our error bounds in Corollary~\ref{cor:SDP_rate}
and Theorem~\ref{thm:cluster_error_rate}.

\textbf{Exact recovery:} If the SNR $\snr$ satisfies the condition~(\ref{eq:snr_cond})
and moreover $\snr^{2}\gtrsim\log\num$, then Theorem~\ref{thm:cluster_error_rate}
guarantees that $\misrate\left(\LabelHat,\LabelStar\right)<\frac{1}{\num}$,
which means that $\misrate\left(\LabelHat,\LabelStar\right)=0$ and
hence the true underlying clusters are recovered exactly. In fact,
in this case Corollary~\ref{cor:SDP_rate} guarantees the SDP solution
satisfies the bound $\norm[\Yhat-\Ystar]1<\frac{1}{4}$, so simply
rounding $\Yhat$ \emph{element-wise }produces the ground-truth cluster
matrix $\Ystar$. Note that the SNR conditions above can be simplified
to $\snr^{2}\gtrsim\numclust+\log\num$ when $\num\gtrsim\vecdim$.

\textbf{Recovery with $\delta$-error:} Our results are applicable
even in the regime with a lower SNR, for which exact recovery of the
clusters is impossible due to the overlap between clusters. In this
regime, Theorem~\ref{thm:cluster_error_rate} implies the following
\emph{approximate recovery} guarantees:\emph{} for any number $\delta\in(0,1)$,
if $\snr$ satisfies the condition~(\ref{eq:snr_cond}) and $\snr^{2}\gtrsim\log\frac{1}{\delta}$,
then $\misrate\left(\LabelHat,\LabelStar\right)\le\delta$ and hence
SDP correctly recovers the cluster memberships of at least $(1-\delta)$
fraction of the points.

In Section \ref{sec:compare} to follow we compare the above results
with existing ones. As a passing note, the above results on \emph{clustering
	error} further imply error bounds on estimating the \emph{cluster
	centers}. We do not delve into the details and refer the readers to
our conference paper \cite{fei2018hidden} for such a result.

\subsection{Robustness under Semi-random Model\label{sec:semirandom}}

In this section we extend the above results to the so called semi-random
setting, in which an omniscient adversary is allowed to modify, in
a coordinated way, the data generated from a probabilistic model.
In the literature \cite{feige2001semirandom,makarychev2012approximation,krivelevich2006coloring,moitra2016robust,awasthi2017clustering},
semi-random models have been recognized as a more flexible model for
non-ideal, real-world data, and are used as a benchmark for evaluating
algorithmic robustness. Algorithms that over-exploit the idealized
structures in a purely random model (e.g., independence, homogeneity
or Gaussianity), often fail completely in the semi-random version
of the model \cite{feige2001semirandom,krivelevich2006coloring,moitra2016robust}.

In the context of SGMM, we consider the following semi-random model
proposed in~\cite{awasthi2017clustering}.

\begin{mdl}[Semi-random Sub-Gaussian Mixture Model]\label{mdl:SGMM_semi}
	
	Let $\{\h_{i}\}$ be points generated from Model~\ref{mdl:SGMM}.
	An adversary can arbitrarily move each point $\h_{i}$ towards its
	cluster center $\Mean_{\labelstar(i)}$ to produce a new point $\widetilde{\h}_{i}$;
	that is
	\[
	\widetilde{\h}_{i}=\Mean_{\labelstar(i)}+\widetilde{\g}_{i},\qquad\text{where }\text{\ensuremath{\widetilde{\g}_{i}=\alpha_{i}\g_{i}} for some \ensuremath{\alpha_{i}}\ensuremath{\in[0,1]}}.
	\]
	Here, $\{\alpha_{i}\}$ are chosen arbitrarily from $[0,1]$ and may
	be correlated with $\{\g_{i}\}$ and with each other.
	
\end{mdl}

Our goal is to estimate the true clustering $\LabelStar$ using the
SDP~(\ref{eq:SDP1}) with the modified data $\{\widetilde{\h}_{i}\}$
as the input. Before proceeding, we make two remarks on the above
model. (i) Although the adversary shrinks the noise $\{\g_{i}\}$,
it does not necessarily make the clustering problem easier, as the
adversary can calibrate its output to create spurious local structures
in the data. (ii) The adversary must move the original point $\h_{i}$
\emph{in the direction of} $\g_{i}=\h_{i}-\Mean_{\labelstar(i)}$.
If the adversary were allowed to move in other directions (without
increasing the distance $\norm[\h_{i}-\Mean_{\labelstar(i)}]2$),
then the clustering structure may be completely lost, since most points
$\widetilde{\h}_{i}$ may become closer to a different cluster center
than to its own center $\Mean_{\labelstar(i)}$. See \cite[Section 1]{awasthi2017clustering}
for a detailed discussion of these two points.\\

Under Model \ref{mdl:SGMM_semi}, we consider the same Oracle IP (\ref{eq:oracleIP})
but with $\{\Mean_{\labelstar(j)}+8\widetilde{\g}_{j}\}$ (that is,
$\{\widetilde{\h}_{i}\}$ with variance augmented by $8$) as the
input. That is, the objective function $\eta$ therein is replaced
by 
\[
\widetilde{\eta}(\F)\coloneqq\sum_{j\in[\num]}\sum_{a\in[\numclust]}\norm[\big(\Mean_{\labelstar(j)}+8\widetilde{\g}_{j}\big)-\Mean_{a}]2^{2}F_{ja}.
\]
The following theorem is an analogue of Theorem~\ref{thm:ip_sdp}
and bounds the SDP error in terms of the Oracle IP error.
\begin{thm}[IP bounds SDP, semi-random]
	\emph{\label{thm:ip_sdp_semi} }Under Model \ref{mdl:SGMM_semi},
	there exist some universal constants $\consts,C>0$ for which the
	following holds. If the SNR satisfies 
	\begin{equation}
	\snr^{2}\geq\consts\numclust,\label{eq:snr_cond_semi}
	\end{equation}
	and $\num\gtrsim\numclust(\vecdim+\log\num)$, then with probability
	at least $1-4\num^{-1}-2^{-\vecdim}$, we have 
	\begin{equation}
	\frac{\norm[\Yhat-\Ystar]1}{\norm[\Ystar]1}\le2\cdot\max\left\{ \frac{\norm[\F-\F^{*}]1}{\norm[\F^{*}]1}:\widetilde{\eta}(\F)\le\widetilde{\eta}(\F^{*}),\F\in\mathcal{F}\right\} +\varepsilon(\num,\numclust,\vecdim,\snr)\label{eq:ip_sdp_semi}
	\end{equation}
	where 
	\[
	\varepsilon(\num,\numclust,\vecdim,\snr):=\widetilde{O}\left(\frac{\numclust\vecdim}{\num\snr^{4}}+e^{-\snr^{4}/C}\right)
	\]
	and $\widetilde{O}(\cdot)$ hides a multiplicative factor of $\log(\vecdim+\log\num)$.
\end{thm}
The proof is given in Appendix~\ref{sec:proof_ip_sdp_semi} and closely
follows that of Theorem \ref{thm:ip_sdp}. We see that the bound is
similar to non-semi-random setting except for the additional, polynomial
error term $\varepsilon(\num,\numclust,\vecdim,\snr)$. This term
captures the ``global'' effect (cf.\ Section~\ref{sec:global_to_local})
of the adversary, who can make some points from two different clusters
closer to each other and thus increase the clustering error.

The next theorem shows that the Oracle IP obeys the same error bound
as in Theorem~\ref{thm:ip_rate}.
\begin{thm}[Exponential rate of IP, semi-random]
	\label{thm:ip_rate_semi} Under Model~\ref{mdl:SGMM_semi}, there
	exist universal constants $\consts,\constgamma,\conste>0$ for which
	the following holds. If $\snr^{2}\ge\consts\numclust$, then we have
	\[
	\max\left\{ \frac{\norm[\F-\F^{*}]1}{\norm[\F^{*}]1}:\widetilde{\eta}(\F)\le\widetilde{\eta}(\F^{*}),\F\in\mathcal{F}\right\} \le\constgamma\exp\left[-\frac{\snr^{2}}{\conste}\right]
	\]
	with probability at least $1-\frac{3}{2}\num^{-1}$.
\end{thm}
The proof is provided in Section~\ref{sec:proof_ip_rate}. Theorem~\ref{thm:ip_rate_semi}
shows that the adversary has essentially \emph{no }``local effect''
(cf.\ Section~\ref{sec:global_to_local}) and does not change the
error of the Oracle IP. This is intuitive: since the Oracle IP knows
the true cluster centers and the adversary can only move points towards
their centers, the error of the Oracle IP can only improve or stay
the same.

Combining Theorems~\ref{thm:ip_sdp_semi} and~\ref{thm:ip_rate_semi}
gives the following explicit error bound for the SDP relaxation.
\begin{cor}[Exponential rate of SDP, semi-random]
	\label{cor:SDP_rate_semi} Under Model~\ref{mdl:SGMM_semi}, the
	SNR condition~(\ref{eq:snr_cond_semi}) and the sample complexity
	condition $\num\gtrsim\numclust(\vecdim+\log\num)$, there exists
	some universal constant $\conste>0$ such that 
	\[
	\misrate(\LabelHat,\LabelStar)\overset{(i)}{\lesssim}\frac{\norm[\Yhat-\Ystar]1}{\norm[\Ystar]1}\overset{(ii)}{\lesssim}\widetilde{O}\left(\frac{\numclust\vecdim}{\num\snr^{4}}+e^{-\snr^{2}/\conste}\right)\overset{(iii)}{\lesssim}\begin{cases}
	\widetilde{O}\left(\frac{\numclust\vecdim}{\num\snr^{4}}\right), & \text{if \ensuremath{\snr^{2}\gtrsim\log(\frac{\num}{\numclust\vecdim})}},\\
	\widetilde{O}\left(e^{-\snr^{2}/\conste}\right), & \text{otherwise,}
	\end{cases}
	\]
	with probability at least $1-6\num^{-1}-2^{-\vecdim}$.
\end{cor}
%\proof{\textit{Proof.}}
\begin{proof}
	The inequality $(i)$ follows from part 1 of Theorem~\ref{thm:cluster_error_rate}.
	The inequality $(ii)$ holds by combining Theorems~\ref{thm:ip_sdp_semi}
	and~\ref{thm:ip_rate_semi} and noting that $e^{-\snr^{4}/C}\le e^{-\snr^{2}/C}$
	as $\snr\ge1$ by assumption. The inequality $(iii)$ holds since
	$\frac{\numclust\vecdim}{\num\snr^{4}}\ge e^{-\snr^{2}/\conste}\Longleftrightarrow\ensuremath{\snr^{2}\ge c_{0}\log(\frac{\num}{\numclust\vecdim})}$
	for a sufficiently large constant $c_{0}>0$.
\end{proof}
%\Halmos
%\endproof
%
The result in Corollary~\ref{cor:SDP_rate_semi} demonstrates a phase
transition phenomenon for the semi-random model. In the low-SNR regime,
the error bound decays exponentially in the SNR~$ \snr $ and is the same as in the non-semi-random
setting. As mentioned, such an error is unavoidable even in standard
SGMM~\cite{lu2016lloyd}. In the high-SNR regime, the effect of the
adversary becomes dominant, in which case the error decays at a slower,
polynomial rate. Interestingly, a lower bound result is established
in~\cite[Theorem 4.1]{awasthi2017clustering}, which shows that any
$\numclust$-means based algorithm must incur this polynomial error$.$
Therefore, we believe that the bound in Corollary~\ref{cor:SDP_rate_semi}
is unimprovable. To the best of our knowledge, this is the first phase
transition result for semi-random SGMM for any algorithm.

\subsubsection{Comparison}

The best result for Semi-random SGMM is given in~\cite{awasthi2017clustering}.
They analyze the Lloyd's k-means algorithm and show that its output
$\LabelHat_{\text{lloyd}}$ satisfies \cite[Theorem 3.1]{awasthi2017clustering}:
\[
\misrate(\LabelHat_{\text{lloyd}},\LabelStar)=\widetilde{O}\left(\frac{\numclust\vecdim}{\num\snr^{4}}\right),\qquad\text{if }\snr^{2}\gtrsim\min\{\numclust,\vecdim\}\cdot\log\num.
\]
Assuming $\numclust\le\vecdim$, one sees that our result in Corollary~\ref{cor:SDP_rate_semi}
strictly generalizes theirs {\cmt under the setting of equal-sized clusters}. In particular, in the high-SNR regime
with $\snr^{2}\gtrsim\log\num$, both results provide the same polynomial
error bound $\widetilde{O}\big(\numclust\vecdim/(\num\snr^{4})\big)$.
Our result further applies to the low-SNR regime with $\snr^{2}\lesssim\log\num$,
while theirs does not. In fact, it is not clear whether the Lloyd's
algorithm can achieve the same robust error bound as the SDP relaxation
in this more challenging regime.

When $k\ge\vecdim$, our result requires the SNR condition $\snr^{2}\gtrsim\numclust$
whereas their result require $\snr^{2}\gtrsim\vecdim\log\num$, which
are incomparable. We suspect that both conditions can be improved,
though we do not have a formal proof.

\subsection{Stochastic Ball Model\label{sec:ball_model}}

In this section, we illustrate the power of our main theorems by deriving
several new results for the Stochastic Ball Model, formally described
below. This model was introduced in \cite{nellore2015recovery} and
has recently attracted attention in the computer science and mathematical
programming communities \cite{awasthi2015relax,iguchi2017certify,li2017kmeans}.

\begin{mdl}[Stochastic Ball Model]\label{mdl:Ball_model} Under Model
	\ref{mdl:SGMM}, we assume in addition that each $\g_{i}$ is sampled
	from a rotationally invariant distribution supported on the unit $\ell_{2}$
	ball in $\real^{\vecdim}$. 
	
\end{mdl}

Under the above model, each data point $\h_{i}=\Mean_{\labelstar(i)}+\g_{i}$
is sampled from the unit ball around its cluster center $\Mean_{\labelstar(i)}$.
It is not hard to see that this model is a special case of SGMM, with
its sub-Gaussian norm given below:
\begin{fact}
	\label{fact:ball_model_sgnorm}Under Model \ref{mdl:Ball_model},
	each $\g_{i}$ has sub-Gaussian norm $\norm[\g_{i}]{\psi_{2}}\le\sgnorm=C\sqrt{\frac{1}{\vecdim}}$
	for some universal constant $C>0$.
\end{fact}
For completeness we prove this standard fact in Appendix~\ref{sec:proof_ball_model_sgnorm}.
Specializing Corollary~\ref{cor:SDP_rate} and Theorem~\ref{thm:cluster_error_rate}
to the Stochastic Ball Model, we obtain the following sufficient conditions
on the minimum center separation $\minsep$ for exact and approximate
recovery:
\[
\minsep^{2}\gtrsim\sqrt{\frac{\numclust}{\num\vecdim}\log\num}+\frac{\numclust}{\num}+\frac{\numclust}{\vecdim}+\begin{cases}
\frac{\log\num}{\vecdim}, & \text{for exact recovery},\\
\frac{\log\delta^{-1}}{\vecdim}, & \text{for recovery of }(1-\delta)\text{ fraction of the points.}
\end{cases}
\]

The state-of-the-art results for Stochastic Ball Models are given
in \cite{awasthi2015relax,iguchi2017certify,li2017kmeans}, which
establish that SDP achieves exact recovery when $\num$ is sufficiently
large and $\minsep^{2}\ge4+\theta(\numclust,\vecdim)$ for some non-negative
function $\theta(\cdot)$. Regardless of the values of $\numclust$
and $\vecdim$, these results all require the separation to be at
least a constant and thus the balls to be disjoint. In contrast, our
results above are applicable to a \emph{small-separation} regime that
is not covered by these existing results. In particular, when $\num$
is large and $\numclust=O(1)$, our results guarantee that SDP achieves
approximate recovery when $\minsep^{2}\gtrsim\frac{1}{\vecdim}$,
which can be arbitrarily smaller than a constant when the dimension
$\vecdim$ grows. Moreover, the recovery is exact if $\minsep^{2}\gtrsim\frac{\log\num}{\vecdim}$,
which can again be arbitrarily small as long as $\num$ does not grow
exponentially fast (i.e., $\num=e^{o(\vecdim)}$). Therefore, in the
high dimensional setting, our results guarantee strong performance
of the SDP\emph{ even when the centers are very close and the balls
	overlap with each other.}

It may appear a bit counter-intuitive that exact/approximate recovery
is achievable when the separation is so small. Such a result is a
manifestation of the geometry in high dimensions: the relative volume
of the intersection of two balls vanishes as the dimension grows.
As a passing note, the result in~\cite[Corollary 4.3]{li2017kmeans}
establishes a \emph{necessary} condition $\minsep\ge1+\sqrt{1+\frac{2}{\vecdim+2}}$
for exact recovery. Our result above does \emph{not} contradict this
condition, as the latter allows $\num$ to grow arbitrarily fast,
in which case with high probability some points will land in the intersection.

\subsection{Comparison with Existing Results\label{sec:compare}}

In this section we compare our results above with the ones in the
literature on clustering SGMM. Our focus is on results that provide
explicit clustering error bounds in the regime where exact cluster
recovery is impossible.

\subsubsection{Clustering Error Bounds}

The work of \cite{mixon2017clustering} considers the Peng-Wei SDP
relaxation introduced in \cite{peng2007approximating}. An intermediate
result of theirs, after appropriate rescaling, establishes the polynomial
error bound $\norm[\Yhat-\Ystar]F^{2}\lesssim\frac{\num^{2}}{\snr^{2}}$
under the setting of balanced clusters and $\snr^{2}\gtrsim\numclust$.
In comparison, our exponential error bound $\norm[\Yhat-\Ystar]F^{2}\le\norm[\Yhat-\Ystar]1\lesssim\frac{\num^{2}}{\numclust}e^{-\snr^{2}}$
is strictly better. The work in \cite{lu2016lloyd} proves an exponentially
decaying error rate similar to ours, but for a different algorithm
(Lloyd's algorithm). Their results require the SNR condition $\snr^{2}\gg\numclust^{2}+\numclust^{3}\frac{\vecdim}{\num}$
and $\numclust^{3}\ll\frac{\num}{\log\num}$ as $\num\to\infty$.
Our SNR condition in (\ref{eq:snr_cond}) has milder dependency on
$\numclust$, though dependency on $\num$ and $\vecdim$ are a bit
more subtle. We do note that under their more restricted SNR condition,
their results provide tight constants in the exponent of the error
rate.

Several papers \cite{giraud2019partial,chen2018hanson,ndaoud2018sharp}
appeared around or after the conference version \cite{fei2018hidden}
of this manuscript was published. All these papers establish an exponential
error bound of the form $\misrate\left(\LabelHat,\LabelStar\right)\lesssim\exp[-\snr^{2}/C]$
that is similar to ours, though the details differ. Restricting to
the spherical Gaussian mixture model with $\numclust=2$ component,
the work~\cite{ndaoud2018sharp} provides a refinement of the above
exponent error bound with a precise coefficient $C$. In particular,
they establish a sharp bound $\misrate\left(\LabelHat,\LabelStar\right)\le\Phi^{c}[(1\pm o(1))r]$,
where $\Phi^{c}$ is the complementary cdf of the standard Gaussian
distribution and $r\coloneqq\snr^{2}/\sqrt{\snr^{2}+\vecdim/\num}$
is a form of SNR. This bound is achieved using an iterative algorithm
that is different from our SDP approach. The work~\cite{chen2018hanson}
generalizes the Peng-Wei SDP relaxation to the non-Euclidean setting.
When specialized to the (Euclidean) SGMM, their error bound is essentially
identical to ours.  The work~\cite{giraud2019partial} also considers
the Peng-Wei SDP relaxation and allows for imbalanced clusters. Their
bound involves an alternative definition for the SNR $\snr_{0}^{2}\coloneqq\min(\snr^{2},\size\snr^{4}R^{-1})$,
where $R$ represents the effective rank of the mixture model.\footnote{Explicitly, $R\coloneqq\frac{\max_{a\in[\numclust]}\norm[\CovMat_{a}]{\text{F}}^{2}}{\max_{a\in[\numclust]}\opnorm{\CovMat_{a}}^{2}}$,
	where $\CovMat_{a}$ is the covariance matrix of the $a$-th cluster.} Since $\snr_{0}\le\snr$, their exponential error bound decays no
faster than ours in Corollary~\ref{cor:SDP_rate}. On the other hand,
their bound holds under the SNR condition $\snr^{2}\gtrsim(1+\sqrt{R/\num})\numclust$,
which is weaker than ours when the effective rank is low. In terms
of the algorithm, they propose to extract explicit clustering from
SDP solutions using an approximate $k$-medoids algorithm. This algorithm
itself involves solving a linear program and is more complicated than
our greedy extraction procedure in Algorithm~\ref{alg:clustering}.

We note that none of the above work provides results for the semi-random
model and the Stochastic Ball Model, unlike what we do in Sections~\ref{sec:semirandom}
and~\ref{sec:ball_model}.

\subsubsection{Conditions for Exact Recovery}
\label{sec:cond_exact_recovery}

As discussed in Section~\ref{sec:consequences}, a corollary of our
results provides sufficient condition, in terms of the SNR $\snr^{2}$,
for exact recovery of the cluster. While exact recovery is not our
focus, we nevertheless summarize and compare with several most representative
results in Table~\ref{tab:compare}. It can be seen that our result
is comparable with, and sometimes better than, the other results in
the table. Only the paper \cite{vempala2004spectral} proves a strictly
better SNR condition, which remains the best to date. Their result
is, however, only established for the special case of mixture of spherical
Gaussians; in fact, their analysis makes substantial use of the structure
of this special case. In comparison, our result is more general and
applies to non-spherical and sub-Gaussian mixtures. Their result also
requires a higher sample complexity $\num=\Omega(\numclust^{2}\vecdim^{3})$
as compared to ours $\num=\Omega(\numclust\vecdim)$.\\

\begin{table}
	\begin{centering}
		\begin{tabular}{|c|c|c|}
			\hline 
			Paper & SNR Condition on $\snr^{2}$ & Algorithm\tabularnewline
			\hline 
			\hline 
			\cite{vempala2004spectral} & %
			\begin{tabular}{c}
				$\Omega\left(\sqrt{\numclust\log\num}\right)$\tabularnewline
				for spherical Gaussian mixture\tabularnewline
			\end{tabular} & Spectral\tabularnewline
			\hline 
			\cite{achlioptas2005spectral} & $\Omega\left(\numclust\log\num+\numclust^{2}\right)$ & Spectral\tabularnewline
			\hline 
			\cite{kumar2010clustering} & $\Omega\left(\numclust^{2}\cdot\text{polylog}\left(\num\right)\right)$ & Spectral\tabularnewline
			\hline 
			\cite{awasthi2012improved} & $\Omega\left(\numclust\cdot\text{polylog}\left(\num\right)\right)$ & Spectral\tabularnewline
			\hline 
			\multirow{1}{*}{\cite{lu2016lloyd}} & $\Omega\left(\numclust^{2}+\log\num\right)$ & \multirow{1}{*}{Spectral + Lloyd's}\tabularnewline
			\hline 
			\cite{li2017kmeans,royer2017adaptive,chen2018hanson,giraud2019partial} & $\Omega\left(\numclust+\log\num\right)$ & SDP\tabularnewline
			\hline 
			\multirow{1}{*}{This paper} & $\Omega\left(\numclust+\log\num\right)$ & \multirow{1}{*}{SDP}\tabularnewline
			\hline 
		\end{tabular}
		\par\end{centering}
	\caption{Summary of existing results on exact cluster recovery for GMM. \label{tab:compare}}
\end{table}

To conclude this section, we mention that there is a large body of
work on estimating the cluster \emph{centers} of Gaussian mixture
model, or estimating the density of the entire mixture distribution.
See, for example, the work in \cite{diakonikolas2017statistical,hopkins2018mixture,kalai2010efficiently,kothari2017better,moitra2010settling,regev2017learning}.
Our work instead focuses on the problem of \emph{clustering} the data
points. Results for these two problems are not directly comparable
in general. There exist settings in which center estimation can be
done while clustering is impossible; for example, when the cluster
centers are identical.

\section{Proof of Theorem \ref{thm:ip_sdp}\label{sec:proof_ip_sdp}}

In this section, we prove Theorem~\ref{thm:ip_sdp}, which relates
the errors of SDP~(\ref{eq:SDP1}) and Oracle IP formulation~(\ref{eq:oracleIP}).
Some additional notations are used in the proof and the rest of this
paper. We define the shorthand $\error\coloneqq\norm[\Yhat-\Ystar]1$
for the $\ell_{1}$ error of the SDP solution. For a matrix $\M$,
we write $\norm[\M]{\infty}\coloneqq\max_{i,j}\left|M_{ij}\right|$
as its entry-wise $\ell_{\infty}$ norm.  We let $\I$ and $\OneMat$
be the $\num\times\num$ identity matrix and all-one matrix, respectively.
For a real number $x$, $\left\lceil x\right\rceil $ denotes its
ceiling. We denote by $\clustset a\coloneqq\left\{ i\in[\num]:\labelstar(i)=a\right\} $
the set of indices of points in cluster $a$, and we define $\size\coloneqq\left|\clustset a\right|=\frac{\num}{\numclust}$
to be the cluster size. Throughout the proof, we use $i,j\in[\num]$
to index the data points, and $a,b\in[\numclust]$ to index the clusters.
We sometimes omit the ranges of these variables, i.e., $[\num]$ and
$[\numclust]$, to avoid cluttered notation.

Our proof consists of three main steps:
\begin{itemize}
	\item \textbf{Step~1:} Use optimality of $\Yhat$ to derive a \emph{basic
		inequality} satisfied by the error matrix $\Yhat-\Ystar$;
	\item \textbf{Step~2:} Relate the basic inequality to a linear program
	(LP) parameterized by the SDP error $\error\coloneqq\norm[\Yhat-\Ystar]1$;
	\item \textbf{Step~3:} Show that the LP is integral and thereby bound $\error$
	by the error of the Oracle IP.
\end{itemize}
Without loss of generality, assume that the SDP error $\error>0$.
For ease of understanding the proof, it is convenient to think of
the $\num$ data points as appropriately ordered and hence $\Ystar$
takes the block diagonal form as in (\ref{eq:block_diagonal}), though
the proof does not actually rely on this ordering.

\subsection*{Step 1: basic inequality}

To streamline the proof, we first record several basic structural
properties of the error matrix $\Yhat-\Ystar$ in the following proposition,
whose proof is given in Section \ref{sec:proof_Yhat-Ystar}.
\begin{fact}
	\label{fact:Yhat-Ystar}Define the shorthand $\B\coloneqq\Yhat-\Ystar$.
	We have
	\begin{enumerate}[label={(\alph*)},ref={\thefact(\alph*)}]
		\item \label{fact:Yhat-Ystar_zero_diag}(zero diagonal) $\forall i\in[\num]:B_{ii}=0$;
		\item \label{fact:Yhat-Ystar_entry_bound}(signs of diagonal and off-diagonal
		blocks) $B_{ij}\in\begin{cases}
		[-1,0], & \text{if }\labelstar_{i}=\labelstar_{j},\\{}
		[0,1], & \text{otherwise};
		\end{cases}$
		\item \label{fact:Yhat-Ystar_zero_row_sum} (zero row sum) $\forall i\in[\num]:\sum_{j:\labelstar(j)\neq\labelstar(i)}B_{ij}=-\sum_{j:\labelstar(j)=\labelstar(i)}B_{ij}$;
		\item \label{fact:Yhat-Ystar_zero_row_sum_block}(zero block row sum) $\forall a\in[\numclust]:\sum_{i:\labelstar(i)=a}\sum_{j:\labelstar(j)\neq a}B_{ij}=-\sum_{i:\labelstar(i)=a}\sum_{j:\labelstar(j)=a}B_{ij}$;
		\item \label{fact:Yhat-Ystar_decompose_gamma}(blockwise decomposition of
		SDP error) $\error=\sum_{i,j:\labelstar(i)\ne\labelstar(j)}B_{ij}-\sum_{i,j:\labelstar(i)=\labelstar(j)}B_{ij}$
		;
		\item \label{fact:Yhat-Ystar_equal_divide_gamma}(diagonal and off-diagonal
		blocks equally divide SDP error) $\frac{\error}{2}=\sum_{i,j:\labelstar(i)\ne\labelstar(j)}B_{ij}=-\sum_{i,j:\labelstar(i)=\labelstar(j)}B_{ij}$.
	\end{enumerate}
\end{fact}
We next decompose the input matrix of pairwise squared distances as
\[
\Adj=\C+\C\t-2\H\H\t,
\]
where $\H\in\real^{\num\times\vecdim}$ is the matrix whose $i$-th
row is the data point $\h_{i}$, and $\C\in\real^{\num\times\num}$
is the matrix where the entries in the $i$-th row are identical and
equal to $\norm[\h_{i}]2^{2}$. By Fact \ref{fact:Yhat-Ystar_zero_row_sum},
we have $\langle\Yhat-\Ystar,\C\rangle=\langle\Yhat-\Ystar,\C\t\rangle=0$,
which implies $\langle\Yhat-\Ystar,\C+\C\t\rangle=0$.

Let $\G\coloneqq\H-\E\H$ be the centered version of $\H$. We can
compute
\[
\H\H\t=\left(\G+\E\H\right)\left(\G+\E\H\right)\t=\G\G\t+\G\left(\E\H\right)\t+\left(\E\H\right)\G\t+\left(\E\H\right)\left(\E\H\right)\t
\]
and 
\[
\E\H\H^{\top}=\E\G\G\t+\left(\E\H\right)\left(\E\H\right)\t.
\]
Therefore, we have the expression
\[
\H\H\t-\E\H\H^{\top}=\left(\G\G\t-\E\G\G\t\right)+\G\left(\E\H\right)\t+\left(\E\H\right)\G\t.
\]
{\cmt Let $\U\coloneqq\frac{1}{\sqrt{\size}}\F^{*}$ so that it takes the form 
%Let $\U\in\real^{\num\times\numclust}$ be the matrix of the left
%singular vectors of $\Ystar$. Note that $\U$ is simply a scaled
%version of the true assignment matrix $\F^{*}$ and takes the form
\[
U_{ia}=\frac{1}{\sqrt{\size}}F_{ia}^{*}=\frac{1}{\sqrt{\size}}\cdot\indic\left\{ \labelstar(i)=a\right\} .
\]
Note that the columns of $\U$ are the left
singular vectors of $\Ystar$.}
For each matrix $\M\in\real^{\num\times\num}$, define the projection
$\PT\left(\M\right)\coloneqq\U\U\t\M+\M\U\U\t-\U\U\t\M\U\U\t$ and
its orthogonal complement $\PTperp(\M)\coloneqq\M-\PT(\M)$.

Now, recall that $\Yhat$ is optimal and $\Ystar$ is feasible to
the SDP~(\ref{eq:SDP1}). Combining with the above decomposition
of $\Adj$ and $\H$, we obtain the following basic inequality:
\begin{align}
0 & \leq-\frac{1}{2}\left\langle \Yhat-\Ystar,\Adj\right\rangle \nonumber \\
& =\left\langle \Yhat-\Ystar,\H\H\t-\E\H\H^{\top}\right\rangle +\left\langle \Yhat-\Ystar,\E\H\H^{\top}\right\rangle \nonumber \\
& =\left\langle \Yhat-\Ystar,\G\G\t-\E\G\G\t+\G\left(\E\H\right)\t+\left(\E\H\right)\G\t\right\rangle +\left\langle \Yhat-\Ystar,\E\H\H^{\top}\right\rangle \nonumber \\
& \overset{(i)}{=}\left\langle \Yhat-\Ystar,\G\G\t\right\rangle +\left\langle \Yhat-\Ystar,\G\left(\E\H\right)\t+\left(\E\H\right)\G\t\right\rangle +\left\langle \Yhat-\Ystar,\left(\E\H\right)\left(\E\H\right)\t\right\rangle \nonumber \\
& =\underbrace{\left\langle \Yhat-\Ystar,\PT\left(\G\G\t\right)\right\rangle }_{S_{1}}+\underbrace{\left\langle \Yhat-\Ystar,\PTperp\left(\G\G\t\right)\right\rangle }_{S_{2}}+2\cdot\underbrace{\left\langle \Yhat-\Ystar,\G\left(\E\H\right)\t\right\rangle }_{S_{3}} \nonumber\\
&\qquad +\underbrace{\left\langle \Yhat-\Ystar,\left(\E\H\right)\left(\E\H\right)\t\right\rangle }_{S_{4}},\label{eq:basic_inequality}
\end{align}
where in step $(i)$ we use the identities $\left\langle \Yhat-\Ystar,\E\G\G\t\right\rangle =0$
and $\left\langle \Yhat-\Ystar,\E\H\H^{\top}\right\rangle =\left\langle \Yhat-\Ystar,\left(\E\H\right)\left(\E\H\right)\t\right\rangle $,
which follow from Fact \ref{fact:Yhat-Ystar_zero_diag} and the fact
that $\E\G\G\t$ is a diagonal matrix.
{\cmt Here, $\Yhat-\Ystar$ takes the role of the error matrix, and $S_1$ and $S_2$ are the products of $\G\G\t$ with the error matrix projected to the column space of $\Ystar$ and its orthogonal complement, respectively. The quantities $S_3$ and $S_4$ can be equivalently written as $S_3 = \sum_{i,j} (\Yhat-\Ystar)_{ij} \langle \g_i, \Mean_{\labelstar(j)} \rangle$ 
and 
$S_4 = \sum_{i,j} (\Yhat-\Ystar)_{ij} \langle \Mean_{\labelstar(i)}, \Mean_{\labelstar(j)} \rangle$.}

\subsection*{Step 2: from SDP to LP}

The following propositions provide high probability bounds on the
terms $S_{1}$, $S_{2}$ and $S_{4}$ that appear on the RHS of the
basic inequality (\ref{eq:basic_inequality}).
\begin{prop}
	\label{prop:S1} If $\snr^{2}\geq C\left(\sqrt{\frac{\numclust\vecdim}{\num}\log\num}+\sqrt{\frac{\numclust}{\num}}\log\num+\frac{\numclust\vecdim}{\num}\right)$
	for some universal constant $C>0$, then $S_{1}\leq\frac{1}{100}\minsep^{2}\error$
	with probability at least $1-\num^{-8}$.
\end{prop}
\begin{prop}
	\label{prop:S2} We have $S_{2}\le\frac{\error}{\size}\opnorm{\G}^{2}$.
	Moreover, if $\snr^{2}\geq C\numclust\left(\frac{\vecdim}{\num}+1\right)$
	for some universal constant $C>0$, then $\frac{\error}{\size}\opnorm{\G}^{2}\leq\frac{1}{100}\minsep^{2}\error$
	with probability at least $1-e^{-\num/2}$.
\end{prop}
\begin{restatable}{prop}{propSfour}\label{prop:S4}
	
	We have $S_{4}=-\frac{1}{2}\sum_{a,b\in[\numclust]:a\ne b}\minsep_{ab}^{2}[\sum_{i\in\clustset a,j\in\clustset b}(\Yhat-\Ystar)_{ij}]$.
	Furthermore, $S_{4}\le-\frac{1}{4}\minsep^{2}\error$.
	
\end{restatable}

We prove the above propositions in Sections \ref{sec:proof_S1}, \ref{sec:proof_S2}
and \ref{sec:proof_S4}, respectively, using appropriate concentration
inequalities. The above bounds on $S_{1},S_{2}$ and $S_{4}$ together
imply that $S_{1}+S_{2}\le-\frac{1}{2}S_{4}$ with probability at
least $1-\left(2\num\right)^{-C'}-2e^{-\num}$ for some universal
constant $C'>0$. Combining with the basic inequality~(\ref{eq:basic_inequality}),
we obtain that with the same probability, there holds
\begin{equation}
0\leq S_{3}+\frac{1}{4}S_{4}.\label{eq:error_S3_bound}
\end{equation}

%The rest of the proof is completely deterministic, in which we exploit
%the above inequality (\ref{eq:error_S3_bound}) and the structures
%of $S_{3}$ and $S_{4}$. 
The rest of the proof is completely deterministic, in which we exploit
the above inequality (\ref{eq:error_S3_bound}) and the structures
of $S_{3}$ and $S_{4}$ {\cmt while treating $\{\g_j\}$ therein as fixed quantities}. 
For each $j\in[\num]$ and $a\in[\numclust]$,
define the variable
\begin{equation}
\beta_{ja}\coloneqq\left\langle \Mean_{a}-\Mean_{\labelstar(j)},\g_{j}\right\rangle -\kappa\minsep_{\labelstar(j),a}^{2},\label{eq:beta_ja}
\end{equation}
where we recall that $\kappa:=1/8$. It is not hard to show that the
quantity $S_{3}+\frac{1}{4}S_{4}$ can be expressed a sum of $\{\beta_{ja}\}$
weighted by entries of the error matrix $\Yhat-\Ystar$. This is the
content of the next lemma, whose proof is given in Section~\ref{sec:proof_S3_S4_identity}.
\begin{lem}
	\label{lem:S3_S4_identity}We have that 
	\begin{equation}
	\frac{1}{\size}\left(S_{3}+\frac{1}{4}S_{4}\right)=\sum_{j}\sum_{a\ne\labelstar(j)}\beta_{ja}\left(\frac{1}{\size}\sum_{i\in\clustset a}\left(\Yhat-\Ystar\right)_{ji}\right).\label{eq:S3_S4}
	\end{equation}
\end{lem}
Note that the RHS of Equation (\ref{eq:S3_S4}) is linear in the quantities
$\widehat{X}_{ja}:=\frac{1}{\size}\sum_{i\in\clustset a}\big(\Yhat-\Ystar\big)_{ji},(j,a)\in[\num]\times[\numclust]$.
With this observation in mind, we proceed to control the RHS of~(\ref{eq:S3_S4})
with a linear program (LP). In particular, consider the following
LP parameterized by a number $R\in[0,\num]$:
\begin{equation}
V(R):=\left\{ \begin{aligned}\max_{\X}\  & \sum_{j}\sum_{a\ne\labelstar(j)}\beta_{ja}X_{ja}\\
\text{s.t.}\  & 0\leq X_{ja}\leq1,\qquad\forall j\in\left[\num\right],a\ne\labelstar(j)\\
& \sum_{a\ne\labelstar(j)}X_{ja}\leq1,\qquad\forall j\in\left[\num\right]\\
& \sum_{j}\sum_{a\ne\labelstar(j)}X_{ja}=R,
\end{aligned}
\right\} .\label{eq:LP}
\end{equation}
Note that this LP is always feasible. Now, Facts~\ref{fact:Yhat-Ystar_entry_bound}
and~\ref{fact:Yhat-Ystar_equal_divide_gamma} imply that 
\[
\frac{\error}{2}=-\sum_{i,j:\labelstar(i)=\labelstar(j)}\left(\Yhat-\Ystar\right)_{ij}\in(0,\num\size]
\]
and thus 
\[
\sum_{j\in\left[\num\right]}\sum_{a\ne\labelstar(j)}\left(\frac{1}{\size}\sum_{i\in\clustset a}\left(\Yhat-\Ystar\right)_{ji}\right)=\frac{1}{\size}\sum_{i,j:\labelstar(i)\ne\labelstar(j)}\left(\Yhat-\Ystar\right)_{ij}=\frac{\error}{2\size}\in(0,\num].
\]
Together with Facts~\ref{fact:Yhat-Ystar_entry_bound} and~\ref{fact:Yhat-Ystar_zero_row_sum},
we conclude that the variables $\widehat{X}_{ja}:=\frac{1}{\size}\sum_{i\in\clustset a}\big(\Yhat-\Ystar\big)_{ji},(j,a)\in[\num]\times[\numclust]$
are feasible to the LP (\ref{eq:LP}) with $R=\frac{\error}{2\size}$,
hence the RHS of (\ref{eq:S3_S4}) is upper bounded by $V\left(\frac{\error}{2\size}\right)$.
Combining with Equation~(\ref{eq:error_S3_bound}), we obtain the
inequality
\begin{equation}
0\leq\frac{1}{\size}\left(S_{3}+\frac{1}{4}S_{4}\right)\leq V\left(\frac{\error}{2\size}\right),\label{eq:basic_ineq_upper_bound_V}
\end{equation}

\subsection*{Step 3: from LP to Oracle IP}

The inequality~(\ref{eq:basic_ineq_upper_bound_V}) immediately implies
a bound on the SDP error: $\frac{\error}{2\size}\le\max\left\{ R\in(0,\num]:V(R)\ge0\right\} $.
It is not hard to see that the last RHS is itself an LP. In fact,
up to a factor of $2$, we can restrict to integer values for the
variable $R$. Indeed, inspecting the LP~(\ref{eq:LP}) we see that
it satisfies $V\left(\frac{\error}{2\size}\right)\le\max\left\{ V\left(\left\lceil \frac{\error}{2\size}\right\rceil \right),V\left(\left\lceil \frac{\error}{2\size}\right\rceil -1\right)\right\} .$
Combining with Equation~(\ref{eq:basic_ineq_upper_bound_V}), we
obtain the bound
\begin{align}
\frac{\error}{2\size}\le\left\lceil \frac{\error}{2\size}\right\rceil  & \le\max\left\{ R\in\{1,2,\ldots,\num\}:V(R)\vee V(R-1)\ge0\right\} \nonumber \\
& \le1+\max\left\{ R\in\{1,2,\ldots,\num\}:V(R)\ge0\right\} \nonumber \\
& \le2\max\left\{ R\in\{1,2,\ldots,\num\}:V(R)\ge0\right\} \nonumber \\
& =2\max\left\{ R\in\{0,1,\ldots,\num\}:V(R)\ge0\right\} .\label{eq:LP_bound}
\end{align}

As the next and crucial step, we observe that for each integer $R\in\{0,1,\ldots,\num\}$,
the LP~(\ref{eq:LP}) defining $V(R)$ is in fact \emph{integral},
that is, it has an optimal solution $\{X_{ja}\}$ satisfying $X_{ja}\in\{0,1\},\forall(j,a)\in[\num]\times[\numclust]$.
Therefore, the value of $V(R)$ is unchanged if we replace the constraint
$0\leq X_{ja}\leq1$ in the LP~(\ref{eq:LP}) with the integer constraint
$X_{ja}\in\{0,1\}$. With this replacement, we can expand the RHS
of the bound~(\ref{eq:LP_bound}) into a single IP: 
\begin{align}
& \max\left\{ R\in\{0,1,\ldots,\num\}:V(R)\ge0\right\} \nonumber \\
= & \left\{ \begin{aligned}\max_{R,\X}\; & R\\
\text{s.t.}\; & R\in\{0,1,\ldots,\num\}\\
& \sum_{j}\sum_{a\ne\labelstar(j)}\beta_{ja}X_{ja}\ge0\\
& X_{ja}\in\{0,1\},\quad\forall j,a\ne\labelstar(j)\\
& \sum_{a\ne\labelstar(j)}X_{ja}\leq1,\quad\forall j\\
& \sum_{j}\sum_{a\ne\labelstar(j)}X_{ja}=R,
\end{aligned}
\right\} =\left\{ \begin{aligned}\max_{\X}\; & \sum_{j}\sum_{a\ne\labelstar(j)}X_{ja}\\
\text{s.t.}\; & \sum_{j}\sum_{a\ne\labelstar(j)}\beta_{ja}X_{ja}\ge0\\
& X_{ja}\in\{0,1\},\quad\forall j,a\ne\labelstar(j)\\
& \sum_{a\ne\labelstar(j)}X_{ja}\leq1,\quad\forall j
\end{aligned}
\right\} =:\IP_{1}.\label{eq:error_bound2}
\end{align}

Note that the above argument is what underlies the hidden integrality
property of the SDP (\ref{eq:SDP1}): even though the \emph{solution}
of the SDP is not integral, the \emph{error} of the solution can be
bounded by an LP that is integral.

As the last step, we show that the $\IP_{1}$ in~(\ref{eq:error_bound2})
above is in fact equivalent to the Oracle IP error as defined in Equation~(\ref{eq:IPerror}).
In particular, recalling that $\eta(\F)$ and $\mathcal{F}$ are the
objective function and feasible set of the Oracle IP, we have the
following identity:
\begin{lem}
	\label{lem:IP_equivalence}We have that 
	\begin{align*}
	\IP_{1} & =\max\left\{ \frac{1}{2}\norm[\F-\F^{*}]1:\eta(\F)\le\eta(\F^{*}),\F\in\mathcal{F}\right\} .
	\end{align*}
\end{lem}
We prove this lemma in Section~\ref{sec:proof_IP_equivalence} by
a careful change-of-variable argument. Combining Equations~(\ref{eq:LP_bound}),
(\ref{eq:error_bound2}) and Lemma~\ref{lem:IP_equivalence}, we
obtain that
\[
\error\le4\size\cdot\IP_{1}=2\size\cdot\max\left\{ \norm[\F-\F^{*}]1:\eta(\F)\le\eta(\F^{*}),\F\in\mathcal{F}\right\} .
\]
Dividing both sides by $\norm[\Ystar]1=\num\size=\norm[\F^{*}]1\size$
proves Theorem \ref{thm:ip_sdp}.

\subsection{Proof of Fact \ref{fact:Yhat-Ystar}\label{sec:proof_Yhat-Ystar}}

Fact \ref{fact:Yhat-Ystar_zero_diag} follows from the fact that $\Yhat$
obeys the diagonal constraint in the SDP (\ref{eq:SDP1}).

Fact \ref{fact:Yhat-Ystar_entry_bound} follows from the fact that
$\Yhat$ is feasible to the SDP (\ref{eq:SDP1}) and hence satisfies
$\yhat_{ij}\in[-1,1]$.

Fact \ref{fact:Yhat-Ystar_zero_row_sum} follows from the fact that
both $\Yhat$ and $\Ystar$ obey the row-sum constraint of the SDP
(\ref{eq:SDP1}).

Fact \ref{fact:Yhat-Ystar_zero_row_sum_block} follows from summing
over $\{i:\labelstar(i)=a\}$ for both sides of the equation in Fact
\ref{fact:Yhat-Ystar_zero_row_sum}.

Fact \ref{fact:Yhat-Ystar_decompose_gamma} follows from Fact \ref{fact:Yhat-Ystar_entry_bound}
and the definition of $\error$.

Fact \ref{fact:Yhat-Ystar_equal_divide_gamma} follows from Facts
\ref{fact:Yhat-Ystar_zero_row_sum_block} and \ref{fact:Yhat-Ystar_decompose_gamma}.

\subsection{Proof of Proposition \ref{prop:S1} \label{sec:proof_S1}}

To bound $S_{1}$, we begin with the decomposition
\begin{align*}
S_{1} & =\left\langle \Yhat-\Ystar,\U\U\t\left(\G\G\t\right)\right\rangle +\left\langle \Yhat-\Ystar,\left(\G\G\t\right)\U\U\t\right\rangle -\left\langle \Yhat-\Ystar,\U\U\t\left(\G\G\t\right)\U\U\t\right\rangle \\
& \leq2\cdot\underbrace{\left|\left\langle \Yhat-\Ystar,\U\U\t\left(\G\G\t\right)\right\rangle \right|}_{T_{1}}+\underbrace{\left|\left\langle \Yhat-\Ystar,\U\U\t\left(\G\G\t\right)\U\U\t\right\rangle \right|}_{T_{2}}.
\end{align*}
By the generalized Holder's inequality, we have the bounds
\begin{align*}
T_{1} & \leq\error\cdot\norm[\U\U\t\left(\G\G\t\right)]{\infty}
\end{align*}
and 
\begin{align*}
T_{2} & =\left|\left\langle \Yhat-\Ystar,\U\U\t\left(\G\G\t\right)\U\U\t\right\rangle \right|\\
& =\left|\left\langle \left(\Yhat-\Ystar\right)\U\U\t,\U\U\t\left(\G\G\t\right)\right\rangle \right|\\
& \leq\error\cdot\norm[\U\U\t\left(\G\G\t\right)]{\infty},
\end{align*}
where the last inequality holds since {\cmt the definition $\U = \frac{1}{\sqrt{\size}} \F^{*}$ implies that $\U\U\t = \frac{1}{\size} \Ystar$, which further  leads to $\norm[\left(\Yhat-\Ystar\right)\U\U\t]1\leq\norm[\Yhat-\Ystar]1=\error$}.
Combining the above bounds, we obtain that
\[
S_{1}\leq3\error\cdot\norm[\U\U\t\left(\G\G\t\right)]{\infty}.
\]

Note that there are $m=\num\numclust$ distinct random variables in
$\U\U\t\left(\G\G\t\right)$ and let us call them $X_{1},\ldots,X_{m}$.
For each $i\in[m]$, one can see that $X_{i}$ takes the form $\left\langle \g_{u(i)},\frac{1}{\left|\calM_{i}\right|}\sum_{j\in\calM_{i}}\g_{j}\right\rangle $
for some index $u(i)\in\left[\num\right]$ and some set $\calM_{i}\subset\left[\num\right]$
with cardinality $\left|\calM_{i}\right|=\size$. Applying the concentration
inequality in Lemma \ref{lem:prod_of_noise_with_sum_noise} with $\calM=\calM_{i}$,
we obtain that for a fixed $i\in[m]$, with probability at least $1-\num^{-10}$
there holds 
\[
X_{i}\le C\sgnorm^{2}\left(\sqrt{\frac{\vecdim\log\num}{\size}}+\frac{\log\num}{\sqrt{\size}}+\frac{\vecdim}{\size}\right)
\]
for some universal constant $C>0$. Applying a union bound gives that
\[
S_{1}\leq3C\error\sgnorm^{2}\left(\sqrt{\frac{\numclust\vecdim\log\num}{\num}}+\sqrt{\frac{\numclust}{\num}}\log\num+\frac{\numclust\vecdim}{\num}\right)
\]
with probability at least $1-\num^{-8}$. The result follows from
the condition of the proposition.

\subsection{Proof of Proposition \ref{prop:S2} \label{sec:proof_S2}}

In this section we control the term $S_{2}$. Since the matrix $\PTperp(\Yhat-\Ystar)=\PTperp(\Yhat)=(\IdMat-\U\U\t)\Yhat(\IdMat-\U\U\t)$
is positive semidefinite, we can apply the generalized Holder's inequality
to obtain
\[
S_{2}=\left\langle \PTperp\left(\Yhat-\Ystar\right),\G\G\t\right\rangle \leq\Tr\left[\PTperp\left(\Yhat-\Ystar\right)\right]\cdot\opnorm{\G\G\t}.
\]
For the second RHS term above, we have the equality $\opnorm{\G\G\t}=\opnorm{\G}^{2}$,
which can be established by taking the Singular Value Decomposition
(SVD) of $\G$. For the first RHS term, we have the identity
\begin{align*}
\Tr\left[\PTperp\left(\Yhat-\Ystar\right)\right] & =\Tr\left[\left(\IdMat-\U\U\t\right)\left(\Yhat-\Ystar\right)\left(\IdMat-\U\U\t\right)\right]\\
& \overset{(i)}{=}\Tr\left[\left(\IdMat-\U\U\t\right)\left(\Yhat-\Ystar\right)\right]\\
& \overset{(ii)}{=}\Tr\left[-\U\U\t\left(\Yhat-\Ystar\right)\right]\\
& =\left\langle -\frac{1}{\size}\Ystar,\Yhat-\Ystar\right\rangle \\
& \overset{(iii)}{=}\frac{\error}{2\size},
\end{align*}
where step $(i)$ holds since the trace is invariant under cyclic
permutation and the matrix $\IdMat-\U\U\t$ is idempotent, step $(ii)$
holds by Fact \ref{fact:Yhat-Ystar_zero_diag}, and step $(iii)$
follows from Fact \ref{fact:Yhat-Ystar_equal_divide_gamma}. Combining
the above results proves that $S_{2}\le\frac{\error}{\size}\opnorm{\G}^{2}$,
the first inequality in the proposition.

We further note the identity $\opnorm{\G}^{2}=\opnorm{\G^{\top}\G}$,
which again follows from the SVD of $\G$. The spectral norm of $\G\t\G=\sum_{i\in\left[\num\right]}\g_{i}\g_{i}^{\top}$
can be controlled using the following standard result.
\begin{lem}[Lemma A.2 in \cite{lu2016lloyd}]
	\label{lem:opnorm_subg_chaos} We have $\opnorm{\sum_{i=1}^{\num}\g_{i}\g_{i}^{\top}}\le6\sgnorm^{2}\left(\num+\vecdim\right)$
	with probability at least $1-e^{-\num/2}$.
\end{lem}
Applying Lemma \ref{lem:opnorm_subg_chaos}, we obtain that with probability
at least $1-e^{-\num/2}$, there holds the inequality
\[
S_{2}\le\frac{\error}{\size}\cdot6\sgnorm^{2}\left(\num+\vecdim\right)=6\error\sgnorm^{2}\numclust\left(1+\frac{\vecdim}{\num}\right).
\]
The second part of the proposition then follows from the assumption
that $\snr^{2}\geq C\numclust\left(\frac{\vecdim}{\num}+1\right)$.

\subsection{Proof of Proposition \ref{prop:S4} \label{sec:proof_S4}}

It can be seen that 
\[
\left(\left(\E\H\right)\left(\E\H\right)\t\right)_{ij}=\begin{cases}
\norm[\Mean_{\labelstar(i)}]2^{2} & \text{if }\labelstar(i)=\labelstar(j),\\
\left\langle \Mean_{\labelstar(i)},\Mean_{\labelstar(j)}\right\rangle  & \text{otherwise}.
\end{cases}
\]
In view of Equation~(\ref{eq:block_diagonal}), we may partition
the matrix $\Yhat-\Ystar$ into $\numclust^{2}$ blocks of size $\size\times\size$.
Define $T_{ab}\coloneqq\sum_{i\in\clustset a,j\in\clustset b}\left(\Yhat-\Ystar\right)_{ij}$
to be the sum of entries within the $(a,b)$-th block. We have 
\begin{align*}
S_{4} & =\sum_{a\in\left[\numclust\right]}T_{aa}\norm[\Mean_{a}]2^{2}+2\sum_{a,b\in\left[\numclust\right]:a<b}T_{ab}\left\langle \Mean_{a},\Mean_{b}\right\rangle \\
& \overset{(i)}{=}-\sum_{a,b\in\left[\numclust\right]:a<b}T_{ab}\minsep_{ab}^{2}\\
& \overset{(ii)}{=}-\frac{1}{2}\sum_{a,b\in\left[\numclust\right]:a\ne b}T_{ab}\minsep_{ab}^{2},
\end{align*}
where step $(i)$ follows from Fact \ref{fact:Yhat-Ystar_zero_row_sum_block},
and step $(ii)$ holds since $T_{ab}=T_{ba}$ as implied by the symmetry
of $\Yhat-\Ystar$. This proves the first part of the proposition.
The second part of the proposition, $S_{4}\le-\frac{1}{4}\minsep^{2}\error$,
follows from combining Fact \ref{fact:Yhat-Ystar_equal_divide_gamma},
the fact that $\minsep_{ab}\ge\minsep$ for $a\ne b$ by definition,
and the fact that $T_{ab}\ge0$ for $a\ne b$ as implied by Fact \ref{fact:Yhat-Ystar_entry_bound}.

\subsection{Proof of Lemma \ref{lem:S3_S4_identity}\label{sec:proof_S3_S4_identity}}

Let $\B\coloneqq\Yhat-\Ystar$. Recall the definitions of $S_{3}$
and $S_{4}$ from Equation (\ref{eq:basic_inequality}). We have 
\begin{align*}
S_{3} & =\sum_{j}\sum_{a}\sum_{i\in C_{a}}B_{ji}\left\langle \Mean_{a},\g_{j}\right\rangle \\
& =\size\sum_{j}\sum_{a}\left\langle \Mean_{a},\g_{j}\right\rangle \left(\frac{1}{\size}\sum_{i\in\clustset a}B_{ji}\right)\\
& =\size\sum_{j}\sum_{a\ne\labelstar(j)}\left\langle \Mean_{a}-\Mean_{\labelstar(j)},\g_{j}\right\rangle \left(\frac{1}{\size}\sum_{i\in\clustset a}B_{ji}\right),
\end{align*}
where the last step holds by Fact \ref{fact:Yhat-Ystar_zero_row_sum}.
On the other hand, by Proposition \ref{prop:S4} we have
\begin{align*}
S_{4} & =-\size\sum_{j}\sum_{a\ne\labelstar(j)}\frac{1}{2}\minsep_{\labelstar(j),a}^{2}\left(\frac{1}{\size}\sum_{i\in\clustset a}B_{ji}\right).
\end{align*}
Summing up the last two equations gives 
\[
\frac{1}{\size}\left(S_{3}+\frac{1}{4}S_{4}\right)=\sum_{j}\sum_{a\ne\labelstar(j)}\left(\left\langle \Mean_{a}-\Mean_{\labelstar(j)},\g_{j}\right\rangle -\frac{1}{8}\minsep_{\labelstar(j),a}^{2}\right)\left(\frac{1}{\size}\sum_{i\in\clustset a}B_{ji}\right).
\]
Note that the quantity inside the first parenthesis above is exactly
$\beta_{ja}$ by definition in Equation~(\ref{eq:beta_ja}). This
completes the proof of the lemma.

\subsection{Proof of Lemma \ref{lem:IP_equivalence}\label{sec:proof_IP_equivalence}}

Recall that $\eta(\F)$ defined in Equation~(\ref{eq:eta_def})
is the objective of the Oracle IP, the set $\mathcal{F}\coloneqq\big\{\F\in\{0,1\}^{\num\times\numclust}:\F\one_{\numclust}=\one_{\num}\big\}$
contains all assignment matrices feasible to the Oracle IP, and $\F^{*}\in\mathcal{F}$
is the true assignment matrix, that is, $F_{ja}^{*}=\indic\left\{ a=\labelstar(j)\right\} $
for all $j\in[\num],a\in[\numclust]$.

Let us reparameterize the integer program $\IP_{1}$ in (\ref{eq:error_bound2})
by a change of variable. Consider any feasible solution $\X\in\{0,1\}^{\num\times\numclust}$
of $\IP_{1}$. For each $j\in[\num]$, we may fix $X_{j,\labelstar(j)}=-\sum_{a\neq\labelstar(j)}X_{ja}$.
Doing so does not affect the feasibility and objective value of $\X$
with respect to $\IP_{1}$. Define the new variable $\F\coloneqq\F^{*}+\X$.
The objective function and constraints of the original variable $\X$
can be mapped to those of $\F$. In particular, for the objective
we have the identity
\begin{align*}
\sum_{j}\sum_{a\ne\labelstar(j)}X_{ja} & =\sum_{j}\sum_{a\ne\labelstar(j)}(F_{ja}-F_{ja}^{*})=\frac{1}{2}\norm[\F-\F^{*}]1.
\end{align*}
For the constraints we have the equivalency
\begin{align*}
\left.\begin{array}{lc}
X_{ja}\in\{0,1\},\quad\forall j\in\left[\num\right],a\ne\labelstar(j)\\
\sum_{a\ne\labelstar(j)}X_{ja}\leq1,\quad\forall j\in\left[\num\right]\\
X_{j,\labelstar(j)}=-\sum_{a\neq\labelstar(j)}X_{ja},\quad\forall j\in[\num]
\end{array}\right\}  & \Longleftrightarrow\F\in\mathcal{F};
\end{align*}
and
\[
\sum_{j}\sum_{a\ne\labelstar(j)}\beta_{ja}X_{ja}\overset{(i)}{=}\sum_{j}\sum_{a}\beta_{ja}X_{ja}=\sum_{j}\sum_{a}\beta_{ja}F_{ja}-\sum_{j}\sum_{a}\beta_{ja}F_{ja}^{*}\overset{(ii)}{=}\sum_{j}\sum_{a}\beta_{ja}F_{ja},
\]
where steps $(i)$ and $(ii)$ both follow from the fact that $\beta_{j,\labelstar(j)}=0,\forall j.$
It follows that $\IP_{1}$ has the same optimal value as another integer
program with the variable $\F$:
\begin{equation}
\IP_{1}=\max\bigg\{\frac{1}{2}\norm[\F-\F^{*}]1:\sum_{j}\sum_{a}\beta_{ja}F_{ja}\ge0,\F\in\mathcal{F}\bigg\}:=\IP_{2}.\label{eq:IP_equivalence}
\end{equation}

We simplify the first constraint in $\IP_{2}$. Recall that $\minsep_{\labelstar(j),a}^{2}=\norm[\Mean_{\labelstar(j)}-\Mean_{a}]2^{2}$
is the separation between clusters $\labelstar(j)$ and $a$, and
$\bar{\h}_{j}\coloneqq\Mean_{\labelstar(j)}+(2\kappa)^{-1}\g_{j},j\in[\num]$
are data points generated from SGMM with augmented variance. By definition
of $\beta_{ja}$, we have 
\begin{align*}
\beta_{ja} & =\left\langle \Mean_{a}-\Mean_{\labelstar(j)},\g_{j}\right\rangle -\kappa\minsep_{\labelstar(j),a}^{2}\\
& =\kappa\left(-\norm[\bar{\h}_{j}-\Mean_{a}]2^{2}+\norm[(2\kappa)^{-1}\g_{j}]2^{2}\right),
\end{align*}
where the last step can be verified by plugging the expressions for
$\minsep_{\labelstar(j),a}^{2}$ and $\bar{\h}_{j}$. It follows
that for each $\F\in\mathcal{F}$, we have
\begin{align*}
\sum_{j}\sum_{a}\beta_{ja}F_{ja} & =\kappa\sum_{j}\sum_{a}\left(-\norm[\bar{\h}_{j}-\Mean_{a}]2^{2}+\norm[(2\kappa)^{-1}\g_{j}]2^{2}\right)F_{ja}\\
& \overset{(i)}{=}\kappa\left(-\sum_{j}\sum_{a}\norm[\bar{\h}_{j}-\Mean_{a}]2^{2}F_{ja}+\sum_{j}\norm[(2\kappa)^{-1}\g_{j}]2^{2}\sum_{a}F_{ja}^{*}\right)\\
& =\kappa\left(-\sum_{j}\sum_{a}\norm[\bar{\h}_{j}-\Mean_{a}]2^{2}F_{ja}+\sum_{j}\sum_{a}\norm[\bar{\h}_{j}-\Mean_{\labelstar(j)}]2^{2}F_{ja}^{*}\right)\\
& \overset{(ii)}{=}\kappa\left(-\sum_{j}\sum_{a}\norm[\bar{\h}_{j}-\Mean_{a}]2^{2}F_{ja}+\sum_{j}\sum_{a}\norm[\bar{\h}_{j}-\Mean_{a}]2^{2}F_{ja}^{*}\right),
\end{align*}
where step $(i)$ holds because $\sum_{a}F_{ja}=1=\sum_{a}F_{ja}^{*},\forall j$,
and step $(ii)$ holds because $F_{ja}^{*}\neq0$ only if $a=\labelstar(j)$.
Recalling the definition $\eta(\F)\coloneqq\sum_{j}\sum_{a}\norm[\bar{\h}_{j}-\Mean_{a}]2^{2}F_{ja}$
given in Section~\ref{sec:oracleIP}, we have the compact expression
\begin{equation}
\sum_{j}\sum_{a}\beta_{ja}F_{ja}=\kappa\left(\eta(\F^{*})-\eta(\F)\right),\quad\forall\F\in\mathcal{F}.\label{eq:eta_func_equivalence}
\end{equation}
Therefore, the first constraint in $\IP_{2}$ is satisfied if and
only if $\eta(\F)\le\eta(\F^{*}).$ Substituting into Equation~(\ref{eq:IP_equivalence}),
we complete the proof of the lemma.

\section{Proof of Theorems \ref{thm:ip_rate} and \ref{thm:ip_rate_semi}\label{sec:proof_ip_rate}}

We first prove Theorem \ref{thm:ip_rate}. The proof of Theorem \ref{thm:ip_rate_semi}
follows from a simple extension.

We define the following shorthand for the error of the Oracle IP:
\[
\widehat{\delta}\coloneqq\max\left\{ \frac{1}{2}\norm[\F-\F^{*}]1:\eta(\F)\le\eta(\F^{*}),\F\in\mathcal{F}\right\} .
\]
Note that $\widehat{\delta}$ takes integer values in $[0,\num]$.
If $\widehat{\delta}=0$ then the theorem follows trivially. We therefore
assume that $\widehat{\delta}\in\{1,2,\ldots,\num\}$. Let $\widehat{\F}\in\{0,1\}^{\num\times\numclust}$
be an optimal solution to the above maximization problem. Define the
matrix $\widehat{\M}\in\{0,1\}^{\num\times\numclust}$ via $\widehat{M}_{ja}:=\widehat{F}_{ja}(1-F_{ja}^{*})$.
We have 
\begin{align}
0 & \le\kappa\left(\eta(\F^{*})-\eta(\widehat{\F})\right)\nonumber \\
& \overset{(i)}{=}\sum_{j\in[\num]}\sum_{a\in[\numclust]}\beta_{ja}\widehat{F}_{ja}\overset{(ii)}{=}\sum_{j\in[\num]}\sum_{a\in[\numclust]}\beta_{ja}\widehat{M}_{ja},\label{eq:ip_rate_proof}
\end{align}
where step $(i)$ holds due to the expression~(\ref{eq:eta_func_equivalence})
of $\eta(\cdot)$, and step $(ii)$ holds since $\widehat{F}_{ja}=\widehat{M}_{ja}$
if $a\neq\labelstar(j)$ and $\beta_{ja}=0$ if $a=\labelstar(j)$.
Note that the random variable $\beta_{ja}$ defined in~(\ref{eq:beta_ja})
has mean $-\kappa\minsep_{\labelstar(j),a}^{2}$ and sub-Gaussian
norm $\norm[\beta_{ja}]{\psi_{2}}\le\sgnorm\minsep_{\labelstar(j),a}$,
and that $\{\beta_{ja}\}$ are independent across $j$. To control
the RHS of (\ref{eq:ip_rate_proof}), we make use of the following
uniform concentration result, which is proved in Section~\ref{sec:proof_lem_order_stat}.
\begin{lem}
	\label{lem:order_stats} Let $\Z\in\real^{\num\times\numclust}$ be
	a matrix with independent rows, such that for each $(j,a)\in[\num]\times[\numclust]$,
	$Z_{ja}$ is a zero-mean sub-Gaussian random variable with sub-Gaussian
	norm no larger than $\rho_{ja}$. Then for some universal constant
	$C>0,$ we have with probability at least $1-\frac{1.5}{\num}$, 
	\begin{equation}
	\begin{aligned}\sum_{j}\sum_{a}\left|\text{\ensuremath{Z_{ja}}}\right|M_{ja} & \le C\sqrt{t\left(\sum_{j}\sum_{a}\rho_{ja}^{2}M_{ja}\right)\log\left(3\num\numclust/t\right)},\\
	& \qquad\quad\forall t\in\{1,2,\ldots,\num\};\forall\M\in\left\{ 0,1\right\} ^{\num\times\numclust}:\M\one_{\numclust}\le\one_{\num},\norm[\M]1=t.
	\end{aligned}
	\label{eq:order_stats}
	\end{equation}
\end{lem}
It is easy to verify that $\widehat{\M}\one_{\numclust}\le\one_{\num}$
and $\norm[\widehat{\M}]1=\frac{1}{2}\norm[\widehat{\F}-\F^{*}]1=\widehat{\delta}$.
Therefore, we can apply Lemma \ref{lem:order_stats} with $Z_{ja}=\beta_{ja}+\kappa\minsep_{\labelstar(j),a}^{2}$
and $\rho_{ja}=\sgnorm\minsep_{\labelstar(j),a}$ to bound the RHS
of (\ref{eq:ip_rate_proof}). Doing so gives that with probability
at least $1-\frac{1.5}{\num}$, we have 
\begin{align*}
0 & \le C\sqrt{\widehat{\delta}\sgnorm^{2}\left(\sum_{j}\sum_{a}\minsep_{\labelstar(j),a}^{2}\widehat{M}_{ja}\right)\log\left(3\num\numclust/\widehat{\delta}\right)}-\kappa\sum_{j}\sum_{a}\minsep_{\labelstar(j),a}^{2}\widehat{M}_{ja}.
\end{align*}

Now, for the the sake of deriving a contradiction, assume that $\widehat{\delta}>3\num\numclust e^{-\snr^{2}/C_{0}^{2}}$
for a fixed constant $C_{0}>C/\kappa$. Continuing from the RHS of
the last display equation, we obtain that 
\begin{align*}
0 & \le C\sqrt{\widehat{\delta}\sgnorm^{2}\left(\sum_{j}\sum_{a}\minsep_{\labelstar(j),a}^{2}\widehat{M}_{ja}\right)\frac{\snr^{2}}{C_{0}^{2}}}-\kappa\sum_{j}\sum_{a}\minsep_{\labelstar(j),a}^{2}\widehat{M}_{ja}\\
& \le\left(\frac{C}{C_{0}}-\kappa\right)\cdot\sum_{j}\sum_{a}\minsep_{\labelstar(j),a}^{2}\widehat{M}_{ja},
\end{align*}
where the last step holds since $\sgnorm^{2}\snr^{2}=\minsep^{2}$
and $\minsep^{2}\widehat{\delta}=\minsep^{2}\sum_{j,a}\widehat{M}_{ja}\le\sum_{j,a}\minsep_{\labelstar(j),a}^{2}\widehat{M}_{ja}$
by definition. Since $C_{0}>C/\kappa$ and $\sum_{j}\sum_{a}\minsep_{\labelstar(j),a}^{2}\widehat{M}_{ja}>0$,
the RHS above is negative, which is a contradiction. Therefore, the
previous assumption is false and we must have
\[
\widehat{\delta}\le3\num\numclust e^{-\snr^{2}/C_{0}^{2}}\overset{(i)}{\le}3\num\numclust\cdot\frac{1}{3\numclust}\cdot e^{-\snr^{2}/\left(2C_{0}^{2}\right)}=\num e^{-\snr^{2}/\left(2C_{0}^{2}\right)},
\]
where step $(i)$ holds under the SNR condition $\snr^{2}\gtrsim\numclust$
assumed in Theorem~\ref{thm:ip_rate}. The theorem then follows from
the fact that $\norm[\F^{*}]1=\num$.\\

Now that we have proved Theorem \ref{thm:ip_rate}, the proof of Theorem
\ref{thm:ip_rate_semi} follows exactly the same argument, except
that all instances of $\g_{j},\beta_{ja},\eta$ above are replaced
by their semi-random versions $\widetilde{\g}_{j},\widetilde{\beta}_{ja},\widetilde{\eta}$
(here $\widetilde{\beta}_{ja}$ is equal to $\beta_{ja}$ defined
in (\ref{eq:beta_ja}) but with $\g_{j}$ replaced by $\widetilde{\g}_{j}$),
and accordingly each $Z_{ja}$ is replaced by $\widetilde{Z}_{ja}=\widetilde{\beta}_{ja}+\kappa\minsep_{\labelstar(j),a}^{2}=\langle\Mean_{a}-\Mean_{\labelstar(j)},\widetilde{\g}_{j}\rangle$.
In the proof we use the distribution of the data only when invoking
Lemma~\ref{lem:order_stats}, whose conclusion~(\ref{eq:order_stats})
remains valid for $\{\widetilde{Z}_{ja}\}$. To see this, recall that
$\widetilde{\g}_{j}=\alpha_{j}\g_{j}$ for some $\alpha_{j}\in[0,1]$.
It follows that $\left|\widetilde{Z}_{ja}\right|=\alpha_{j}\left|Z_{ja}\right|\le\left|Z_{ja}\right|$,
so the LHS of (\ref{eq:order_stats}) does not increase.

\subsection{Proof of Lemma \ref{lem:order_stats} \label{sec:proof_lem_order_stat}}

We define the quantities
\begin{align*}
L(\M,\b) & \coloneqq\sum_{j,a}b_{j}Z_{ja}M_{ja},\qquad\text{for each }\M\in\{0,1\}^{\num\times\numclust},\b\in\{\pm1,0\}^{\num},\\
R(\M,t) & \coloneqq C\sqrt{t\left(\sum_{j,a}\rho_{ja}^{2}M_{ja}\right)\log\left(3\num\numclust/t\right)},\qquad\text{for each }\M\in\{0,1\}^{\num\times\numclust},t\in[\num]
\end{align*}
and the sets
\begin{align*}
\calM(t) & \coloneqq\left\{ \M\in\left\{ 0,1\right\} ^{\num\times\numclust}:\M\one_{\numclust}\le\one_{\num},\norm[\M]1=t\right\} ,\qquad\text{for each }t\in[\num],\\
{\cal B}(\M,t) & :=\{\b\in\{\pm1,0\}^{\num}:b_{j}=0\text{ if }M_{ja}=0,\text{ \ensuremath{\forall a\in[\numclust]}}\},\qquad\text{for each }\M\in\calM(t),t\in[\num].
\end{align*}

We begin by bounding the probability
\begin{align*}
\alpha_{t} & \coloneqq\P\bigg\{\exists\M\in\calM(t):\sum_{j,a}\left|Z_{ja}\right|M_{ja}>R(\M,t)\bigg\}
\end{align*}
for each integer $t\in[\num]$. To this end, note that $\big|Z_{ja}\big|=\max_{b_{j}\in\{\pm1\}}\big\{ b_{j}Z_{ja}\big\}$,
hence
\begin{align}
\alpha_{t} & \le\P\Big\{\exists\M\in\calM(t),\exists\b\in\calB(\M,t):L(\M,\b)>R(\M,t)\Big\}\nonumber \\
& \leq\sum_{\M\in\calM(t)}\sum_{\b\in\calB(\M,t)}\P\Big\{ L(\M,\b)>R(\M,t)\Big\}.\label{eq:double union bd on normal}
\end{align}
By assumption, the $Z_{ja}$'s are independent zero-mean sub-Gaussian
random variables, so the squared sub-Gaussian norm of the sum $L(\M,\b)$
is at most $C_{\psi_{2}}\sum_{j,a}\rho_{ja}^{2}M_{ja}$ where $C_{\psi_{2}}>0$
is a universal constant. We can therefore apply Hoeffding's inequality
(Lemma \ref{lem:hoeffding}) to bound each summand on the RHS of (\ref{eq:double union bd on normal}):
\begin{align*}
\P\left\{ L(\M,\b)>R(\M,t)\right\}  & \leq\exp\left\{ -\frac{c_{0}C^{2}t\left(\sum_{j,a}\rho_{ja}^{2}M_{ja}\right)\log(3\num\numclust/t)}{C_{\psi_{2}}\sum_{j,a}\rho_{ja}^{2}M_{ja}}\right\} \\
& \leq\exp\left\{ -4t\log(3\num\numclust/t)\right\} ,
\end{align*}
where $c_{0}>0$ is a universal constant. Plugging this back to (\ref{eq:double union bd on normal}),
we have for each $t\in\left[\num\right]$:
\begin{align*}
\alpha_{t} & \leq\sum_{\M\in\calM(t)}\sum_{\b\in\calB(\M,t)}\exp\left\{ -4t\log(3\num\numclust/t)\right\} \\
& =\binom{\num}{t}\numclust^{t}\cdot2^{t}\cdot\exp\left\{ -4t\log(3\num\numclust/t)\right\} \\
& \leq\left(\frac{\num e}{t}\right)^{t}\numclust^{t}\cdot2^{t}\cdot\exp\left\{ -4t\log(3\num\numclust/t)\right\} \\
& \overset{(i)}{\le}\exp\left\{ 2t\log(3\num\numclust/t)-4t\log(3\num\numclust/t)\right\} \\
& \leq\left(\frac{t}{3\num\numclust}\right)^{t},
\end{align*}
where step $(i)$ follows from $t\leq t\log(3\num\numclust/t)$ for
$t\in\left[\num\right]$ and $\numclust\ge2$.

With the above bound on $\alpha_{t}$ and a union bound over $t\in[\num]$,
we can control the probability that the conclusion of the lemma fails
to hold:
\[
\P\bigg\{\exists t\in[\num],\exists\M\in\calM(t):\sum_{j,a}\left|Z_{ja}\right|M_{ja}>R(\M,t)\bigg\}\le\sum_{t=1}^{\num}\alpha_{t}\leq\sum_{t=1}^{\num}\left(\frac{t}{3\num\numclust}\right)^{t}.
\]
The proof is complete if we can show that the last RHS is at most
$\frac{1.5}{\num}$. Note that 
\[
\sum_{t=1}^{\num}\left(\frac{t}{3\num\numclust}\right)^{t}\le\frac{1}{3\num}+\sum_{t=2}^{\num}\left(\frac{t}{3\num}\right)^{t}\le\frac{1}{3\num}+\num\cdot\max_{t\in[2,\num]}\left(\frac{t}{3\num}\right)^{t}.
\]
Hence it suffices to show that for all $t\in[2,\num]$, there holds
$\left(\frac{t}{3\num}\right)^{t}\leq\frac{1}{\num^{2}}$ or equivalently
$f(t)\coloneqq t(\log3\num-\log t)\geq2\log\num.$ For $t\in[2,\num]$,
the function $f$ has derivative 
\[
f'(t)=\log3\num-\log t-1\ge\log3\num-\log\left(\num\right)-1=\log3-1\ge0.
\]
Therefore, $f(t)$ is non-decreasing for $t\in[2,\num]$ and thus
$f(t)\ge f(2)=2\log3\num-2\log2\ge2\log\num$ as desired. We conclude
that $\sum_{t=1}^{\num}\left(\frac{t}{3\num\numclust}\right)^{t}\le\frac{1.5}{\num}$,
thereby completing the proof of Lemma~\ref{lem:order_stats}.

\section{Conclusion\label{sec:conclusion}}

In this paper, we study the performance of SDP relaxation for clustering
SGMM and its semi-random version. Our analysis proceeds in two steps:
(a) bound the clustering error of the SDP by that of the idealized
Oracle IP; (b) show that the error of the Oracle IP decays exponentially
in the SNR. As mentioned, this two-step framework allows for a decoupling
of the computational and statistical mechanisms that drive the performance
of the SDP approach. The oracle bound in step (a) represents a fundamental
performance limit of the SDP relaxation. We expect that further progress
in understanding SDP relaxations is likely to come from improvements
in this step. On the other hand, step (b) is problem-specific and
makes use of the probabilistic structures of the model. By modifying
and sharpening this step, one may generalize our results to other
variants of SGMM.

Our work points to several interesting future directions. An immediate
problem is to obtain tighter, less pessimistic bounds for mixtures
with inhomogeneous components, which may have different pairwise separation
and variance along different directions. It is also of interest to
study other forms of robustness of SDP relaxations, e.g., in the presence
of arbitrary outliers and model misspecification. Other directions
worth exploring include obtaining better constants in error bounds,
identifying sharp thresholds for different types of recovery, and
developing scalable computational procedures for solving the SDP.
%{\cmt 
%	Last but not least, it would be interesting to understand the relation between parameter estimation and clustering, and investigate whether our separation condition for the SDP approach can be improved or not.
%}

Last but not least, another interesting future direction is to better understand the relation between the clustering problem---which we have focused on---and the related parameter estimation problem (see Section~\ref{sec:intro}). 
%In particular, it is of interest to investigate whether our SNR condition for the SDP approach can be further improved. 
Note that a guarantee for one problem can be converted to a guarantee for the other. While this conversion may not be tight in general, it is nevertheless interesting to compare our SNR condition with those in existing work on both clustering and parameter estimation. The work of \cite{regev2017learning} shows that the SNR condition $\snr^2 \gtrsim \log \numclust$ is sufficient and necessary to achieve a constant error in parameter estimation with polynomial sample complexity. Note that the algorithm in  \cite{regev2017learning} requires an initialization procedure that runs in time exponential in $\numclust $, the number of clusters.
In the special case of Gaussian Mixture Model,  the work \cite{vempala2004spectral} proves that spectral methods achieve exact cluster recovery under the SNR condition $\snr^2 \gtrsim \sqrt{\numclust \log \num}$; see Section~\ref{sec:cond_exact_recovery} for further discussion of the work~\cite{vempala2004spectral}. These  results seem to suggest that our SNR condition $\snr^2 \gtrsim \numclust + \log\num$ may have a suboptimal dependence on $ \numclust $. It is of interest to investigate whether this potential suboptimality is intrinsic to the SDP relaxation approach or can be avoided by a tighter analysis.

%The statistical upper and lower bounds of $\log \numclust$ given by \cite{regev2017learning} for parameter estimation might also suggest that the $\Omega(\numclust)$ dependency in our SNR condition could be unnecessary statistically.

\appendix
\appendixpage
	
\section{Proof of Theorem \ref{thm:cluster_error_rate}\label{sec:proof_cluster_error_rate}}

We only need to prove the first part of the theorem. The second part
follows immediately from the first part and Theorem~\ref{cor:SDP_rate}.

The proof follows a similar strategy as those of \cite[Theorem 17 and Lemma 18]{makarychev2016learning}
and is divided into two lemmas. Recall that Algorithm~\ref{alg:apx_clustering}
outputs a collection of sets $\left\{ B_{t}\right\} _{t\ge1}$ that
are not necessarily equal-sized. The first lemma characterizes the
quality of these sets. Define the shorthands $\numclust':=\left|\left\{ B_{t}\right\} _{t\ge1}\right|$
(which satisfies $\numclust'\ge\numclust$) and $\epsilon\coloneqq\norm[\Yhat-\Ystar]1/\norm[\Ystar]1$.
\begin{lem}
	\label{lem:apx_clustering} There exists a partial matching $\perm'$
	(i.e., an injection from $\left[\numclust\right]$ to $\left[\numclust'\right]$)
	and a universal constant $C>0$ such that 
	\[
	\left|\bigcup_{a\in[\numclust]}\clustset a\cap B_{\pi'(a)}\right|\ge\left(1-C\epsilon\right)\num.
	\]
\end{lem}
The proof is given in Section~\ref{sec:proof_apx_clustering}. Building
on the above lemma, the next lemma further characterizes the (equal-sized)
clusters $\left\{ U_{t}\right\} _{t\in[k]}$ obtained in Algorithm~\ref{alg:clustering}.
\begin{lem}
	\label{lem:clustering} There exists a permutation $\perm$ on $\left[\numclust\right]$
	and a universal constant $C>0$ such that
	\[
	\left|\bigcup_{a\in[\numclust]}\clustset a\cap U_{\perm(a)}\right|\ge\left(1-C\epsilon\right)\num.
	\]
\end{lem}
The proof is given in Section \ref{sec:proof_clustering}. In light
of the last lemma and the fact that 
\[
\misrate(\LabelHat,\LabelStar)=1-\frac{1}{\num}\max_{\perm\in S_{\numclust}}\left|\bigcup_{a\in[\numclust]}\clustset a\cap U_{\perm(a)}\right|,
\]
we obtain the bound $\misrate(\LabelHat,\LabelStar)\le C\epsilon$
as stated in the first part of Theorem~\ref{thm:cluster_error_rate}.

\subsection{Proof of Lemma \ref{lem:apx_clustering}\label{sec:proof_apx_clustering}}

Let $\y_{a}\in\real^{\num}$ be one of the $\size$ identical rows
of $\Ystar$ whose row indices are in $\clustset a$. Define the sets
\begin{align*}
G_{a} & \coloneqq\left\{ i\in\clustset a:\norm[\Yhat_{i\bullet}-\y_{a}]1\leq\frac{\size}{8}\right\} ,\qquad\forall a\in\left[\numclust\right]
\end{align*}
and let $G\coloneqq\bigcup_{a\in\left[\numclust\right]}G_{a}$ and
$H\coloneqq\vertexset\backslash G$, where $V:=[\num]$.

A partial matching between the sets $\{\clustset a\}_{a\in[\numclust]}$
and $\{B_{t}\}_{t\in[\numclust']}$ is given by an injective function
$\perm'$ from $[\numclust]$ to $[\numclust']$. We construct such
a partial matching by matching each cluster $\clustset a$ with the
first $B_{t}$ that intersects $G_{a}$; i.e., we set $\perm'(a)=\min\{t\in[\numclust']:B_{t}\cap G_{a}\neq\emptyset\}.$
Since each $i\in\left[\num\right]$ belongs to some $B_{t}$, every
$\clustset a$ is matched with some $B_{t}$. Moreover, we show below
that $\pi'$ is indeed injective as it cannot match two distinct clusters
$\clustset a$ and $\clustset b$ with the same $B_{t}$.
\begin{claim}
	For each $a\in\left[\numclust\right]$ and $t\in\left[\numclust'\right]$
	such that $t=\perm'(a)$, we have that $B_{t}\cap G_{b}=\emptyset$
	for any $b\in[\numclust]\setminus\{a\}$ and that $B_{t}\subset G_{a}\cup H$.
\end{claim}
%\proof{\textit{Proof.}}
\begin{proof}
	Suppose that there exist $B_{t}$ and $b\in\left[\numclust\right]\setminus\{a\}$
	such that $B_{t}\cap G_{b}\ne\emptyset$. Let $u\in B_{t}\cap G_{a}$
	and $v\in B_{t}\cap G_{b}$. Since $G_{a}$ and $G_{b}$ are disjoint,
	we know that $u\ne v$. Let $w\in B_{t}$. Then we have 
	\[
	\norm[\Yhat_{u\bullet}-\Yhat_{w\bullet}]1\leq\frac{\size}{4}\quad\text{and}\quad\norm[\Yhat_{v\bullet}-\Yhat_{w\bullet}]1\leq\frac{\size}{4},
	\]
	whence 
	\[
	\norm[\Yhat_{u\bullet}-\Yhat_{v\bullet}]1\leq\norm[\Yhat_{u\bullet}-\Yhat_{w\bullet}]1+\norm[\Yhat_{v\bullet}-\Yhat_{w\bullet}]1\le\frac{\size}{2}.
	\]
	This implies that 
	\begin{align*}
	\norm[\y_{a}-\y_{b}]1 & \leq\norm[\y_{a}-\Yhat_{u\bullet}]1+\norm[\Yhat_{u\bullet}-\Yhat_{v\bullet}]1+\norm[\y_{b}-\Yhat_{v\bullet}]1\\
	& \leq\frac{\size}{8}+\frac{\size}{2}+\frac{\size}{8}<\size,
	\end{align*}
	which is a contradiction to the fact that $\norm[\y_{a}-\y_{b}]1=2\size$.
	To complete the proof, we note that for any $i\in B_{t}$ we have
	either $i\in G_{a}$ or $i\in H$, hence $B_{t}\subset G_{a}\cup H.$
\end{proof}
%\Halmos
%\endproof
The rest of the proof proceeds by establishing the following three
claims.
\begin{claim}
	For each $a\in\left[\numclust\right]$ and $t\in\left[\numclust'\right]$
	such that $t=\perm'(a)$, we have 
	\begin{equation}
	\left|B_{t}\cap\clustset a\right|\geq\left|G_{a}\right|-\left|B_{t}\cap H\right|.\label{eq:claim2}
	\end{equation}
\end{claim}
%\proof{\textit{Proof.}}
\begin{proof}
	Fix $i\in G_{a}$ for some $a\in\left[\numclust\right]$. For any
	$j\in G_{a}$ we have $j\in B(i)$ since 
	\[
	\norm[\Yhat_{i\bullet}-\Yhat_{j\bullet}]1\le\norm[\y_{a}-\Yhat_{i\bullet}]1+\norm[\y_{a}-\Yhat_{j\bullet}]1\le\frac{\size}{4}.
	\]
	Therefore, by definition we have $\left|B_{t}\right|\ge\left|B(i)\right|\ge\left|G_{a}\right|.$
	It follows that 
	\begin{align*}
	\left|B_{t}\cap\clustset a\right| & \overset{(i)}{\ge}\left|B_{t}\cap G_{a}\right|\\
	& =\left|B_{t}\right|-\left|B_{t}\backslash G_{a}\right|\\
	& \overset{(ii)}{=}\left|B_{t}\right|-\left|B_{t}\cap H\right|\\
	& \ge\left|G_{a}\right|-\left|B_{t}\cap H\right|,
	\end{align*}
	where step $(i)$ holds since $G_{a}\subset\clustset a$ and step
	$(ii)$ holds since $B_{t}\subset G_{a}\cup H$ by the previous claim.
\end{proof}
%\Halmos
%\endproof
\begin{claim}
	We have 
	\[
	\sum_{(t,a):t=\perm'(a)}\left|B_{t}\cap\clustset a\right|\ge\left|\vertexset\right|-2\left|H\right|.
	\]
\end{claim}
%\proof{\textit{Proof.}}
\begin{proof}
	Summing both sides of Equation~(\ref{eq:claim2}) over $\{(t,a):t=\perm'(a)\}$,
	we obtain
	\begin{align*}
	\sum_{(t,a):t=\perm'(a)}\left|B_{t}\cap\clustset a\right| & =\sum_{a\in\left[\numclust\right]}\left|G_{a}\right|-\sum_{(t,a):t=\perm'(a)}\left|B_{t}\cap H\right|\\
	& \ge\sum_{a\in\left[\numclust\right]}\left|G_{a}\right|-\sum_{t\ge1}\left|B_{t}\cap H\right|\\
	& \overset{(i)}{=}\left|G\right|-\left|\vertexset\cap H\right|\\
	& =\left|\vertexset\right|-2\left|H\right|,
	\end{align*}
	where step $(i)$ holds since the sets $\{B_{t}\cap H\}$ are disjoint
	and $\bigcup_{t\geq1}B_{t}=\vertexset$.
\end{proof}
%\Halmos
%\endproof
\begin{claim}
	There exists a universal constant $C>0$ such that $\left|H\right|\leq C\epsilon\num$.
\end{claim}
%\proof{\textit{Proof.}}
\begin{proof}
	We have 
	\[
	\left|H\right|\cdot\frac{\size}{8}\leq\sum_{i\in H}\norm[\Yhat_{i\bullet}-\y_{\labelstar(i)}]1\leq\norm[\Yhat-\Ystar]1\leq\epsilon\norm[\Ystar]1=\epsilon\cdot\num\size
	\]
	where the last step follows from the fact that $\norm[\Ystar]1=\num\size$.
\end{proof}
%\Halmos
%\endproof
Combining the last two claims proves Lemma~\ref{lem:apx_clustering}.

\subsection{Proof of Lemma \ref{lem:clustering}\label{sec:proof_clustering}}

Without loss of generality, assume that the output of Algorithm~\ref{alg:apx_clustering}
is ordered as $\left|B_{1}\right|\ge\left|B_{2}\right|\ge\cdots\ge\left|B_{\numclust'}\right|$.
Consequently, the sets $\left\{ U_{t}\right\} _{t\in\left[\numclust\right]}$
maintained in Algorithm~\ref{alg:clustering} are such that each
$U_{t}$ consists of $B_{t}$ and some elements from the sets $B_{u}$
with $u>\numclust$.

Let $\perm'$ be the partial matching between $\{\clustset a\}_{a\in[\numclust]}$
and $\{B_{t}\}_{t\in[\numclust']}$ given by Lemma~\ref{lem:apx_clustering}.
Define $\perm(a)=\perm'(a)$ for each $a\in[\numclust]$ with $\perm'(a)\le\numclust$,
and extend $\perm$ to a full permutation on $\left[\numclust\right]$
in an arbitrary way. We have 
\begin{align*}
\left|\bigcup_{(t,a):t=\perm(a)}\clustset a\cap U_{t}\right| & \ge\left|\bigcup_{(t,a):t=\perm'(a)\le\numclust}\clustset a\cap B_{t}\right|\\
& =\left|\bigcup_{(t,a):t=\perm'(a)}\clustset a\cap B_{t}\right|-\left|\bigcup_{(t,a):t=\perm'(a)>\numclust}\clustset a\cap B_{t}\right|\\
& \ge\left(1-C'\epsilon\right)\num-\left|\bigcup_{(t,a):t=\perm'(a)>\numclust}\clustset a\cap B_{t}\right|,
\end{align*}
where the last step follows from Lemma~\ref{lem:apx_clustering}
and $C'>0$ is a universal constant. Define the sets
\begin{align*}
T_{1} & \coloneqq\left\{ t>\numclust:t=\perm'(a)\text{ for some }a\in\left[\numclust\right]\right\} ,\\
T_{2} & \coloneqq\left\{ t\in\left[\numclust\right]:t\ne\perm'(a)\text{ for all }a\in\left[\numclust\right]\right\} .
\end{align*}
It is easy to verify that $\left|T_{1}\right|=\left|T_{2}\right|$
and that $\left|B_{t_{1}}\right|\le\left|B_{t_{2}}\right|$ for each
$t_{1}\in T_{1}$ and $t_{2}\in T_{2}$. It follows that 
\begin{align*}
\left|\bigcup_{(t,a):t=\perm'(a)>\numclust}\clustset a\cap B_{t}\right|\le\left|\bigcup_{t\in T_{1}}B_{t}\right| & \le\left|\bigcup_{t\in T_{2}}B_{t}\right|\\
& \le\left|\vertexset\right|-\left|\bigcup_{(t,a):t=\perm'(a)}\clustset a\cap B_{t}\right|\\
& \le C'\epsilon\num,
\end{align*}
where the last step follows again from Lemma~\ref{lem:apx_clustering}.
Combining pieces and setting $C\coloneqq2C'$, we obtain the desired
bound
\[
\left|\bigcup_{(t,a):t=\perm(a)}\clustset a\cap U_{t}\right|\ge\left(1-C'\epsilon\right)\num-C'\epsilon\num=(1-C\epsilon)\num.
\]

\section{Proof of Theorem \ref{thm:ip_sdp_semi} \label{sec:proof_ip_sdp_semi}}

Let $\widetilde{\G}\in\real^{\num\times\vecdim}$ be the matrix whose
$j$-th row is $\widetilde{\g}_{j}$, and recall that the $j$-th
row of $\E\H$ is $\Mean_{\labelstar(j)}$. Then using the same notations
as in Section \ref{sec:proof_ip_sdp}, we have the following analogue
of Equation (\ref{eq:basic_inequality}):
\begin{align}
0 & \le\underbrace{\left\langle \Yhat-\Ystar,\PT\left(\widetilde{\G}\widetilde{\G}\t\right)\right\rangle }_{S_{1}}+\underbrace{\left\langle \Yhat-\Ystar,\PTperp\left(\widetilde{\G}\widetilde{\G}\t\right)\right\rangle }_{S_{2}}\nonumber \\
& \qquad+2\underbrace{\left\langle \Yhat-\Ystar,\widetilde{\G}\left(\E\H\right)\t\right\rangle }_{S_{3}}+\underbrace{\left\langle \Yhat-\Ystar,\left(\E\H\right)\left(\E\H\right)\t\right\rangle }_{S_{4}}.\label{eq:basic_inequality_semi}
\end{align}
As will become clear shortly, the impact of the semi-random adversary
is essentially limited to the term $S_{1}$. With this effect accounted
for, the proof of Theorem~\ref{thm:ip_sdp_semi} is basically the
same as that of Theorem~\ref{thm:ip_sdp} for the non-semi-random
setting.

We begin by controlling $S_{1}$. This is done in the following proposition,
whose proof is given in Section \ref{sec:proof_prop_S1_semi}. Recall
that $\widetilde{O}(\cdot)$ hides a multiplicative factor of $\log(\vecdim+\log\num)$.
\begin{prop}
	\label{prop:S1_semi} If $\num\gtrsim\numclust(\vecdim+\log\num)$,
	then there exists some universal constant $C>0$ such that 
	\[
	S_{1}\le\frac{1}{100}\error\minsep^{2}+\sgnorm^{2}\cdot\widetilde{O}\left(\frac{\num\vecdim}{\snr^{2}}+\num^{2}\snr^{2}e^{-\snr^{4}/C}\right).
	\]
	with probability at least $1-2\num^{-1}-2^{-\vecdim}$.
\end{prop}
In comparison with its non-semi-random counterpart in Proposition~\ref{prop:S1},
the bound in Proposition~\ref{prop:S1_semi} has an additional error
term that is due to the effect of the adversary.

Controlling the term $S_{2}$ is straightforward. In particular, the
following proposition establishes a bound that is exactly the same
as in the non-semi-random setting (cf.~Proposition \ref{prop:S2}),
since the adversary cannot make the bound worse.
\begin{prop}
	\label{prop:S2_semi} If $\snr^{2}\geq C\numclust\left(\frac{\vecdim}{\num}+1\right)$
	for some universal constant $C>0$, then $S_{2}\leq\frac{1}{100}\minsep^{2}\error$
	with probability at least $1-e^{-\num/2}$.
\end{prop}
%\proof{\textit{Proof.}}
\begin{proof}
	By the first part of Proposition \ref{prop:S2}, we have $S_{2}\le\frac{\error}{\size}\opnorm{\widetilde{\G}}^{2}$.
	Since each row of $\widetilde{\G}$ is a shrunk version of the corresponding
	row of $\G$, we have 
	\[
	\opnorm{\widetilde{\G}}=\max_{\v\in\real^{\vecdim}:\norm[\v]2=1}\norm[\widetilde{\G}\v]2\le\max_{\v\in\real^{\vecdim}:\norm[\v]2=1}\norm[\G\v]2=\opnorm{\G},
	\]
	whence $S_{2}\le\frac{\error}{\size}\opnorm{\G}^{2}$. Applying the
	second part of Proposition \ref{prop:S2} proves the desired bound.
\end{proof}
%\Halmos
%\endproof
Note that the SNR condition in Proposition \ref{prop:S2_semi} is
satisfied by our assumption $\num\gtrsim\numclust\vecdim$ and $\snr^{2}\gtrsim\numclust$.

The term $S_{4}$ is unaffected by the adversary and thus can be bounded
as before using Proposition \ref{prop:S4}, which is re-stated below
for readers' convenience.

\propSfour*

We are now ready to prove Theorem~\ref{thm:ip_sdp_semi}. Propositions~\ref{prop:S2_semi}
and \ref{prop:S4} imply that $S_{2}\le-\frac{1}{2}S_{4}$ with probability
at least $1-e^{-\num/2}\ge1-2\num^{-1}$. Plugging into the basic
inequality (\ref{eq:basic_inequality_semi}), we obtain that $0\le S_{1}+2S_{3}+\frac{1}{2}S_{4},$
hence we must have either (a) $0\le2S_{3}+\frac{1}{4}S_{4}$ or (b)
$0\le S_{1}+\frac{1}{4}S_{4}.$ Let us analyze each of these two cases.

\paragraph{Case (a): $0\le2S_{3}+\frac{1}{4}S_{4}$.}

In this case, we continue the proof in exactly the same way as in
Section~\ref{sec:proof_ip_sdp} following Equation~(\ref{eq:error_S3_bound}),
but replace $\{\g_{j}\}$ and $\eta$ therein with $\{\widetilde{\g}_{j}\}$
and $\widetilde{\eta}$. Doing so establishes the desired inequality
(\ref{eq:ip_sdp_semi}) (without the second RHS term).

\paragraph{Case (b): $0\le S_{1}+\frac{1}{4}S_{4}$.}

Applying Proposition \ref{prop:S1_semi} to bound $S_{1}$ and Proposition
\ref{prop:S4} to bound $S_{4}$, we obtain that with probability
at least $1-2\num^{-1}-2^{-\vecdim}$,
\[
\error\minsep^{2}\le\sgnorm^{2}\cdot\widetilde{O}\left(\frac{\num\vecdim}{\snr^{2}}+\num^{2}\snr^{2}e^{-\snr^{4}/C}\right).
\]
Dividing both sides of the above equation by $\minsep^{2}\norm[\Ystar]1=\minsep^{2}\frac{\num^{2}}{\numclust}$
and recalling that $\snr:=\frac{\minsep}{\sgnorm}$, we get 
\[
\frac{\norm[\Yhat-\Ystar]1}{\norm[\Ystar]1}\le\widetilde{O}\left(\frac{\numclust\vecdim}{\num\snr^{4}}+\numclust e^{-\snr^{4}/C}\right)\le\widetilde{O}\left(\frac{\numclust\vecdim}{\num\snr^{4}}+e^{-\snr^{4}/C_{0}}\right),
\]
where the last step holds since our assumption $\snr^{2}\gtrsim\numclust$
implies that $\numclust e^{-\snr^{4}/C}\le e^{-\snr^{4}/C_{0}}$ for
some universal constant $C_{0}\ge C$. We have therefore established
the desired inequality (\ref{eq:ip_sdp_semi}) (without the first
RHS term).

Combining the above two cases completes the proof of Theorem \ref{thm:ip_sdp_semi}.

\subsection{Proof of Proposition \ref{prop:S1_semi}\label{sec:proof_prop_S1_semi}}

We first state several technical lemmas concerning the vectors $\{\widetilde{\g}_{i}\}$
produced by the semi-random adversary. Introduce the shorthand 
\begin{equation}
\vecdim_{0}:=\vecdim+\log\num,\label{eq:effective_dim}
\end{equation}
which can be interpreted as the effective dimension and is motivated
by the following two lemmas.
\begin{lem}
	\label{lem:semi_subgauss_norm_bound} With probability at least $1-\num^{-1}$,
	for some universal constant $c\ge1$ we have 
	\[
	\norm[\widetilde{\g}_{i}]2\le c\sgnorm\sqrt{\vecdim_{0}}\qquad\text{uniformly for all }i\in[\num].
	\]
\end{lem}
%\proof{\textit{Proof.}}
\begin{proof}
	By definition of $\widetilde{\g}_{i}$ we have $\norm[\widetilde{\g}_{i}]2\le\norm[\g_{i}]2$.
	Standard results on the norms of sub-Gaussian vectors (e.g., \cite[Theorem 3.1.1]{vershynin2017high})
	ensure that $\norm[\g_{i}]2\le\frac{1}{2}c\sgnorm(\sqrt{\vecdim}+\sqrt{\log\num})\le c\sgnorm\sqrt{\vecdim_{0}}$
	with probability at least $1-\num^{-2}$. Taking a union bound over
	all $i\in[\num]$ proves the lemma.
\end{proof}
%\Halmos
%\endproof
\begin{lem}
	\label{lem:semi_average_bound}If $\size\ge4\vecdim_{0}$, then with
	probability at least $1-\num^{-1}$, we have 
	\[
	\norm[\frac{1}{\size}\sum_{i\in C_{a}^{*}}\widetilde{\g}_{i}]2\le3\sgnorm\qquad\text{uniformly for all }a\in[\numclust].
	\]
\end{lem}
%\proof{\textit{Proof.}}
\begin{proof}
	Fix $a\in[\numclust]$. We have the expression $\frac{1}{\size}\sum_{i\in C_{a}^{*}}\widetilde{\g}_{i}=\frac{1}{\size}\B_{a}\D\onevec$,
	where $\B_{a}\in\real^{\vecdim\times\size}$ is the matrix with columns
	$\{\g_{i},i\in C_{a}^{*}\}$, $\D\in\real^{\size\times\size}$ is
	a diagonal matrix with diagonal entries $\big\{\norm[\widetilde{\g}_{i}]2/\norm[\g_{i}]2,i\in C_{a}^{*}\big\}$,
	and $\onevec\in\real^{\size}$ is the all-one vector. It follows that
	$\norm[\frac{1}{\size}\sum_{i\in C_{a}^{*}}\widetilde{\g}_{i}]2\le\frac{1}{\size}\opnorm{\B_{a}}\cdot\opnorm{\D}\cdot\norm[\onevec]2$.
	But $\opnorm{\D}\le1$, $\norm[\onevec]2=\sqrt{\size}$, and Lemma~\ref{lem:opnorm_subg_chaos}
	guarantees that $\opnorm{\B_{a}}=\sqrt{\opnorm{\B_{a}\B_{a}^{\top}}}\le\sqrt{6\sgnorm^{2}(\size+\vecdim)}$
	with probability at least $1-e^{-\size/2}\ge1-\num^{-2}$ (since $\size\ge4\log\num$
	by assumption). Combining pieces, we have $\norm[\frac{1}{\size}\sum_{i\in C_{a}^{*}}\widetilde{\g}_{i}]2\le\frac{1}{\size}\cdot\sqrt{6\sgnorm^{2}(\size+\vecdim)}\cdot\sqrt{\size}\le3\sgnorm$,
	where the last step holds since $\size\ge4\vecdim$ by assumption.
	Taking a union bound over all $a\in[\numclust]$ proves the lemma.
\end{proof}
%\Halmos
%\endproof
To state the next lemma, we define the set $\calB_{\lambda}(\v):=\left\{ i\in[\num]:\left|\left\langle \widetilde{\g}_{i},\v\right\rangle \right|>\lambda\sgnorm\right\} $
for each unit vector $\v$ in $\real^{\vecdim}$ and each positive
real number $\lambda$. Also recall that $\widetilde{O}(\cdot)$ hides
a multiplicative factor of $\log(\vecdim+\log\num)$. The following
key lemma is proved at the end of this section.
\begin{lem}
	\label{lem:bad_direction}Suppose that $\num\ge\vecdim$. There exists
	a universal constant $C>0$ such that with probability at least $1-2^{-\vecdim}-\num^{-1}$,
	uniformly for all numbers $\lambda\ge1$ and all unit vectors $\v\in\real^{\vecdim}$,
	we have 
	\begin{equation}
	\left|\calB_{\lambda}(\v)\right|=\widetilde{O}\left(\frac{\vecdim}{\lambda^{2}}+\num\exp\left(-\frac{\lambda^{2}}{C}\right)\right).\label{eq:bad_direction}
	\end{equation}
\end{lem}
{\cmt A similar result to Lemma \ref{lem:bad_direction} has appeared in  \cite[Lemma 2.11]{awasthi2017clustering}. The main difference between Lemma \ref{lem:bad_direction} and the existing result is that we have the exponential term $\num e^{-\lambda^2 / C}$ in Equation \eqref{eq:bad_direction}. As it will turn out later, the quantity $\lambda$  plays the role of the SNR $\snr$. Therefore, in the low-SNR regime, this exponential term dominates \eqref{eq:bad_direction} and leads to the final exponential rate in Corollary \ref{cor:SDP_rate_semi}.}

We are now ready to prove Proposition \ref{prop:S1_semi}. Note that
the conditions in Lemma~\ref{lem:semi_average_bound} and~\ref{lem:bad_direction}
are satisfied under the assumption of Proposition~\ref{prop:S1_semi}.
Assume that the conclusions of Lemmas \ref{lem:semi_subgauss_norm_bound}\textendash \ref{lem:bad_direction}
hold simultaneously; this event has probability at least $1-2\num^{-1}-2^{-\vecdim}$,
which is the probability claimed in Proposition~\ref{prop:S1_semi}.

We begin by using the definition of $\PT$ to obtain the decomposition
\begin{align*}
\left|\left(\PT\left(\widetilde{\G}\widetilde{\G}\t\right)\right)_{ij}\right| & \le\underbrace{\left|\left\langle \widetilde{\g}_{i},\frac{1}{\size}\sum_{u\in\clustset{\labelstar(j)}}\widetilde{\g}_{u}\right\rangle \right|}_{T_{ij}}+\underbrace{\left|\left\langle \widetilde{\g}_{j},\frac{1}{\size}\sum_{u\in\clustset{\labelstar(i)}}\widetilde{\g}_{u}\right\rangle \right|}_{T_{ji}} \\
& \qquad +\underbrace{\left|\left\langle \frac{1}{\size}\sum_{w\in\clustset{\labelstar(i)}}\widetilde{\g}_{w},\frac{1}{\size}\sum_{u\in\clustset{\labelstar(j)}}\widetilde{\g}_{u}\right\rangle \right|}_{S_{ij}}
\end{align*}
for each $i,j\in[\num]$. It is easy to see that $S_{ij}\le\frac{1}{\size}\sum_{w\in\clustset{\labelstar(i)}}T_{wj}$
by triangle inequality. With these inequalities and introducing the
shorthand $D_{ij}:=\left|(\Yhat-\Ystar)_{ij}\right|$, we can bound
the quantity of interest $S_{1}:=\big\langle\Yhat-\Ystar,\PT(\widetilde{\G}\widetilde{\G}\t)\big\rangle$
as follows:
\begin{align}
S_{1} & \le\sum_{i,j}D_{ij}\cdot(T_{ij}+T_{ji}+S_{ij})\le2\sum_{j}\sum_{i}D_{ij}T_{ij}+\frac{1}{\size}\sum_{j}\sum_{i}\sum_{w\in\clustset{\labelstar(i)}}D_{wj}T_{ij},\label{eq:semi_sum}
\end{align}
where the last step follows from the symmetry of the matrix $\Yhat-\Ystar$
and a change of indexing.

To proceed, we fix an arbitrary $j\in[\num]$. Let $\v_{j}:=\frac{1}{\size}\sum_{u\in\clustset{\labelstar(j)}}\widetilde{\g}_{u}/\norm[\frac{1}{\size}\sum_{u\in\clustset{\labelstar(j)}}\widetilde{\g}_{u}]2$
and note that $\v_{j}$ is a unit vector. Set $\lambda_{0}:=\frac{\minsep^{2}}{C_{0}\sgnorm^{2}}=\frac{\snr^{2}}{C_{0}}$
for some sufficiently large constant $C_{0}>0$, and note that $\lambda_{0}\ge\numclust\ge1$
by our assumption on $\snr$. Recalling the definition of the set
$\calB_{\lambda}(\cdot)$, we consider two cases:
\begin{itemize}
	\item For each $i\in[\num]\backslash\calB_{\lambda_{0}}(\v_{j})$, we have
	\[
	T_{ij}=\left|\left\langle \widetilde{\g}_{i},\v_{j}\right\rangle \right|\cdot\norm[\frac{1}{\size}\sum_{u\in\clustset{\labelstar(j)}}\widetilde{\g}_{u}]2\le\lambda_{0}\sgnorm\cdot4\sgnorm,
	\]
	where in the last step we use the defining property of the set $[\num]\backslash\calB_{j}$
	to bound the first term, and use Lemma~\ref{lem:semi_average_bound}
	to bound the second term.
	\item For $i\in\calB_{\lambda_{0}}(\v_{j})$, we have 
	\begin{align*}
	\sum_{i\in\calB_{\lambda_{0}}(\v_{j})}T_{ij} & \le\sum_{i\in\calB_{\lambda_{0}}(\v_{j})}\left|\left\langle \widetilde{\g}_{i},\v_{j}\right\rangle \right|\cdot4\sgnorm\\
	& =\sum_{i\in\calB_{\lambda_{0}}(\v_{j})}\int_{0}^{\infty}\indic\left\{ \lambda\tau<\left|\left\langle \widetilde{\g}_{i},\v_{j}\right\rangle \right|\right\} \diff\left(\lambda\sgnorm\right)\cdot4\sgnorm\\
	& =4\sgnorm^{2}\int_{0}^{\infty}\sum_{i\in\calB_{\lambda_{0}}(\v_{j})}\indic\left\{ \lambda\tau<\left|\left\langle \widetilde{\g}_{i},\v_{j}\right\rangle \right|\right\} \diff\lambda\\
	& =4\sgnorm^{2}\int_{0}^{\infty}\left|\calB_{\lambda}(\v_{j})\cap\calB_{\lambda_{0}}(\v_{j})\right|\diff\lambda.
	\end{align*}
	Applying Lemma~\ref{lem:bad_direction} to bound the sizes of $\calB_{\lambda}(\v_{j})$
	and $\calB_{\lambda_{0}}(\v_{j})$, we obtain
	\begin{align*}
	\sum_{i\in\calB_{\lambda_{0}}(\v_{j})}T_{ij} & =\widetilde{O}\left(\sgnorm^{2}\int_{0}^{\infty}\min\left\{ \frac{\vecdim}{\lambda^{2}}+\num e^{-\lambda^{2}/C},\frac{\vecdim}{\lambda_{0}^{2}}+\num e^{-\lambda_{0}^{2}/C}\right\} \diff\lambda\right)\\
	& =\sgnorm^{2}\widetilde{O}\left(\int_{\lambda_{0}}^{\infty}\left(\frac{\vecdim}{\lambda^{2}}+\num e^{-\lambda^{2}/C}\right)\diff\lambda+\lambda_{0}\cdot\left(\frac{\vecdim}{\lambda_{0}^{2}}+\num e^{-\lambda_{0}^{2}/C}\right)\right)\\
	& =\sgnorm^{2}\widetilde{O}\left(\frac{\vecdim}{\lambda_{0}}+\num\lambda_{0}e^{-\lambda_{0}^{2}/C}\right),
	\end{align*}
	where $C$ is the universal constant $C$ given in Lemma~\ref{lem:bad_direction}.
\end{itemize}
With the above two bounds and the fact that $D_{ij}\le1,\forall i,j$
(implied by Fact \ref{fact:Yhat-Ystar_entry_bound}), we can bound
the first RHS term in Equation~(\ref{eq:semi_sum}) as 
\begin{align*}
\sum_{j}\sum_{i}D_{ij}T_{ij} & \le\sum_{j}\sum_{i\in[\num]\backslash\calB_{\lambda_{0}}(\v_{j})}D_{ij}T_{ij}+\sum_{j}\sum_{i\in\calB_{\lambda_{0}}(\v_{j})}D_{ij}T_{ij}\\
& \le\left(\sum_{j}\sum_{i\in[\num]\backslash\calB_{\lambda_{0}}(\v_{j})}D_{ij}\right)\cdot4\lambda_{0}\sgnorm^{2}+\sum_{j}\left(\sum_{i\in\calB_{\lambda_{0}}(\v_{j})}T_{ij}\right)\\
& \lesssim\error\cdot\lambda_{0}\sgnorm^{2}+\num\cdot\sgnorm^{2}\widetilde{O}\left(\frac{\vecdim}{\lambda_{0}}+\num\lambda_{0}e^{-\lambda_{0}^{2}/C}\right).
\end{align*}
The second RHS term in Equation~(\ref{eq:semi_sum}) can be controlled
in a similar fashion and obey the same bound. Combining these bounds
and recalling that $\lambda_{0}:=\frac{\minsep^{2}}{C_{0}\sgnorm^{2}}=\frac{\snr^{2}}{C_{0}}$,
we obtain 
\[
S_{1}\lesssim\error\lambda_{0}\sgnorm^{2}+\num\sgnorm^{2}\widetilde{O}\left(\frac{\vecdim}{\lambda_{0}}+\num\lambda_{0}e^{-\lambda_{0}^{2}/C}\right)\le\frac{1}{100}\error\minsep^{2}+\sgnorm^{2}\cdot\widetilde{O}\left(\frac{\num\vecdim}{\snr^{2}}+\num^{2}\snr^{2}e^{-\snr^{4}/C_{1}}\right)
\]
for some universal constant $C_{1}>0$, thereby proving Proposition~\ref{prop:S1_semi}.

\subsubsection{Proof of Lemma \ref{lem:bad_direction}}

Our strategy involves two steps: (i) prove the desired inequality
for fixed $\lambda$ and $\v$, and (ii) establish a uniform bound
using an $\epsilon$-net argument.

\paragraph{Step (i): Controlling $\left|\protect\calB_{\lambda}(\protect\v)\right|$
	for fixed $\lambda$ and $\protect\v$.}

Set $\epsilon:=1/(2c\sqrt{\vecdim_{0}})$, where $c$ is as defined
in Lemma \ref{lem:semi_subgauss_norm_bound} and $\vecdim_{0}$ is
defined in Equation (\ref{eq:effective_dim}). We assume without loss
of generality that $c\sqrt{\vecdim_{0}}$ is an integer; otherwise,
we replace $c$ in the definition of $\epsilon$ with a larger constant
so as to fulfill this assumption. Let us first show that the inequality~(\ref{eq:bad_direction})
holds for a fixed number $\lambda\ge1$ and a unit vector $\v$ with
probability at least $1-(cd_{0})^{-1}(3/\epsilon)^{-2\vecdim}$. Define
the indicator random variable $X_{i}:=\indic\left\{ \left|\left\langle \g_{i},\v\right\rangle \right|>\lambda\sgnorm\right\} $
for $i\in[\num]$. Since each $\widetilde{\g}_{i}$ is a shrunk version
of $\g_{i}$, the quantity of interest satisfies the bound $\left|\calB_{\lambda}(\v)\right|\le\sum_{i\in[\num]}X_{i}$.
The last RHS is the sum of independent Bernoulli RVs, where 
\[
\mu:=\E\left[\sum_{i\in[\num]}X_{i}\right]=\sum_{i\in[\num]}\P\left\{ \left|\left\langle \g_{i},\v\right\rangle \right|>\lambda\sgnorm\right\} \le\num e^{-\lambda^{2}/8}
\]
as $\left\langle \g_{i},\bar{\v}\right\rangle $ has sub-Gaussian
norm at most $\sgnorm$. To bound the sum $\sum_{i}X_{i}$, we record
the standard Chernoff bound:
\begin{lem}[Chernoff bound; Theorem 4.4 in \cite{mitzenmacher2005probability}]
	Under the above setting, for each $\delta>0$ we have $\P\left\{ \sum_{i}X_{i}\ge(1+\delta)\mu\right\} \le e^{-\mu(1+\delta)\log(1+\delta)+\mu\delta}\le e^{-\mu\min\{\delta^{2},\delta\}/3}$
\end{lem}
We proceed by considering two cases:

\emph{Case 1:} If $\vecdim<\num e^{-\lambda^{2}/16}$, then we set
$\delta=\left(\sqrt{\frac{9d}{\mu}}+\frac{9\vecdim}{\mu}\right)\log(\frac{3}{\epsilon})$.
Applying the second inequality in the Chernoff bound, we obtain that
with probability at least $1-\exp\left(-3\vecdim\log(\frac{3}{\epsilon})\right)\ge1-(c\sqrt{d_{0}})^{-1}(3/\epsilon)^{-2\vecdim}$,
there holds
\[
\sum_{i}X_{i}\lesssim\left(\mu+d\right)\log(\frac{3}{\epsilon})=\widetilde{O}\left(\num e^{-\lambda^{2}/8}+\num e^{-\lambda^{2}/16}\right)=\widetilde{O}\left(\num e^{-\lambda^{2}/16}\right),
\]
where we recall that note $\widetilde{O}(\cdot)$ hides multiplicative
factors of $\log(\frac{3}{\epsilon})\asymp\log(\vecdim+\log\num)$.

\emph{Case 2:} If $\vecdim\ge\num e^{-\lambda^{2}/16}$, then we set
$\delta=\left(3+\frac{18d}{\mu}/\log(\frac{18d}{\mu})\right)\log(\frac{3}{\epsilon})$.
In this case, we have the inequalities 
\begin{align*}
\delta\ge3\implies(1+\delta)\log(1+\delta)\ge\frac{3}{2}\delta & \implies(1+\delta)\log(1+\delta)-\delta\ge\frac{1}{3}\delta\log\delta
\end{align*}
and
\begin{align*}
\mu\le\num e^{-\lambda^{2}/8}\le\frac{\vecdim^{2}}{\num}\implies\frac{18\vecdim}{\mu}\ge\frac{18\num}{\vecdim}\ge18 & \implies\log(\frac{18\vecdim}{\mu})-\log\log(\frac{18\vecdim}{\mu})\ge\frac{1}{2}\log(\frac{18\vecdim}{\mu})\ge0,
\end{align*}
whence
\[
\mu(1+\delta)\log(1+\delta)-\mu\delta\ge\frac{1}{3}\mu\delta\cdot\log\delta\ge\frac{6\vecdim\log(\frac{3}{\epsilon})}{\log(\frac{18\vecdim}{\mu})}\cdot\left(\log(\frac{18\vecdim}{\mu})-\log\log(\frac{18\vecdim}{\mu})\right)\ge3\vecdim\log(\frac{3}{\epsilon}).
\]
Applying the first inequality in the Chernoff bound, we obtain that
with probability at least $1-\exp\left(-3\vecdim\log(\frac{3}{\epsilon})\right)\ge1-(c\sqrt{d_{0}})^{-1}(3/\epsilon)^{-2\vecdim}$,
there holds
\[
\sum_{i}X_{i}\lesssim\mu+\frac{\vecdim\log(\frac{3}{\epsilon})}{\log(18\vecdim/\mu)}=\widetilde{O}\left(\mu+\frac{d}{\log(18)+\lambda^{2}/8-\log(n/\vecdim)}\right)=\widetilde{O}\left(\num e^{-\lambda^{2}/8}+\frac{\vecdim}{\lambda^{2}/16}\right),
\]
where the last step holds because $\vecdim\ge\num e^{-\lambda^{2}/16}\implies\log(\num/\vecdim)\le\frac{\lambda^{2}}{16}$.

\paragraph{Step (ii): Applying union bound on $\epsilon$-net and for all $\lambda$.}

In both cases above, the inequality~(\ref{eq:bad_direction}) holds
with probability $\ge1-(c\sqrt{d_{0}})^{-1}(3/\epsilon)^{-2\vecdim}$,
for each fixed number $\lambda\ge1$ and unit vector $\v$. Now, let
$\mathcal{N}$ be an $\epsilon$-net of the set of $\vecdim$-dimensional
unit vectors, where $\left|\mathcal{N}\right|\le(3/\epsilon)^{\vecdim}$.
Applying a union bound, we find that the inequality~(\ref{eq:bad_direction})
holds simultaneously for all integers $\bar{\lambda}=1,2,\ldots,c\sqrt{\vecdim_{0}}$
and vectors $\bar{\v}\in\mathcal{N}$ with probability at least $1-(3/\epsilon)^{-\vecdim}\ge1-6^{-\vecdim}$
(since $\epsilon:=1/(2c\sqrt{\vecdim_{0}})\le1/2$). Also note that
$\left|\calB_{c\sqrt{\vecdim_{0}}}(\v)\right|=0$ with probability
at least $1-\num^{-1}$ (cf.\ Lemma~\ref{lem:semi_subgauss_norm_bound}).

On the above event, for all \emph{real numbers }$\lambda\in[1,c\sqrt{\vecdim_{0}}]$
and all vectors $\bar{\v}\in\mathcal{N}$, we have 
\[
\left|\calB_{\lambda}(\bar{\v})\right|\le\left|\calB_{\left\lfloor \lambda\right\rfloor }(\bar{\v})\right|=\widetilde{O}\left(\frac{\vecdim}{\left\lfloor \lambda\right\rfloor ^{2}}+\num\exp\left(-\frac{\left\lfloor \lambda\right\rfloor ^{2}}{C}\right)\right)\le\widetilde{O}\left(\frac{4\vecdim}{\lambda}+\num\exp\left(-\frac{\lambda^{2}}{4C}\right)\right),
\]
where the last two steps hold since $\left\lfloor \lambda\right\rfloor $
is an integer and satisfy $\left\lfloor \lambda\right\rfloor \ge\lambda/2$.
Moreover, for all $\lambda>c\sqrt{\vecdim_{0}}$ we have $\left|\calB_{\lambda}(\bar{\v})\right|\le\left|\calB_{c\sqrt{\vecdim_{0}}}(\bar{\v})\right|=0$
. We hence see that the inequality~(\ref{eq:bad_direction}) holds
(with a change of absolute constants) for all \emph{$\lambda\ge1$}
and $\bar{\v}\in\mathcal{N}$. Finally, for each unit vector $\v$
in $\real^{\vecdim}$, let $\bar{\v}$ be the nearest vector in the
$\epsilon$-net $\mathcal{N}$. If $i\in\calB_{\lambda}(\v)$, then
\[
\left|\left\langle \widetilde{\g}_{i},\bar{\v}\right\rangle \right|\ge\left|\left\langle \widetilde{\g}_{i},\v\right\rangle \right|-\left|\left\langle \widetilde{\g}_{i},\bar{\v}-\v\right\rangle \right|\overset{(i)}{\ge}\lambda\sgnorm-c\sgnorm\sqrt{\vecdim_{0}}\cdot\epsilon\overset{(ii)}{\ge}\frac{1}{2}\lambda\sgnorm,
\]
where step $(i)$ follows from Lemma~\ref{lem:semi_subgauss_norm_bound}
and step $(ii)$ follows from our choice $\epsilon:=1/(2c\sqrt{\vecdim_{0}})$
and $\lambda\ge1$. This means that $\calB_{\lambda}(\v)\subseteq\calB_{\lambda/2}(\bar{\v})$,
and thus inequality~(\ref{eq:bad_direction}) holds (with a change
of absolute constants) for all unit vectors $\v$ in $\real^{\vecdim}$
as well. We have completed the proof of Lemma~\ref{lem:bad_direction}.

\section{Standard Results for Sub-Gaussian Random Variables}

We collect several standard tail bounds that are used in the proofs
of our main theorems.

\subsection{Sub-Gaussian Tail Bounds}

The first lemma is the standard Hoeffding's inequality as given in
\cite[Theorem 2.6.2]{vershynin2017high}.
\begin{lem}[Hoeffding's inequality for Sub-Gaussians]
	\emph{ \label{lem:hoeffding} }Let $X_{1},\ldots,X_{N}$ be independent,
	mean zero, sub-Gaussian random variables. Then, for every $t\geq0$
	we have 
	\[
	\P\left[\left|\sum_{i=1}^{N}X_{i}\right|\geq t\right]\leq2\exp\left[-\frac{ct^{2}}{\sum_{i=1}^{N}\norm[X_{i}]{\psi_{2}}^{2}}\right],
	\]
	where $c>0$ is a universal constant.
\end{lem}

The next lemma controls the inner product between sub-Gaussian random
vectors.
\begin{lem}
	\label{lem:prod_of_noise_with_sum_noise} Let $\left\{ \x_{i}\right\} _{i\in\left[\num\right]}$
	be independent sub-Gaussian random vectors such that $\norm[\x_{i}]{\psi_{2}}\le\rho$
	for each $i\in\left[\num\right]$. For any fixed $i\in\left[\num\right]$,
	$\calM\subset\left[\num\right]$, $t>0$ and $\delta>0$, there exists
	a universal constant $C>0$ such that 
	\[
	\left\langle \x_{i},\frac{1}{\left|\calM\right|}\sum_{j\in\calM}\x_{j}\right\rangle \le\frac{3\rho^{2}\left(5C\sqrt{\left|\calM\right|}\left(\sqrt{\vecdim\log\num}+\log\num\right)+\vecdim\right)}{\left|\calM\right|}
	\]
	with probability at least $1-\num^{-10}$.
\end{lem}
%\proof{\textit{Proof.}}
\begin{proof}
	We record the following lemma, which is Lemma A.3 in \cite{lu2016lloyd}.
	\begin{lem}
		Let $\left\{ \x_{i}\right\} _{i\in\left[\num\right]}$ be independent
		sub-Gaussian random vectors such that $\norm[\x_{i}]{\psi_{2}}\le\rho$
		for each $i\in\left[\num\right]$. For any fixed $i\in\left[\num\right]$,
		$\calM\subset\left[\num\right]$, $t>0$ and $\delta>0$, we have
		\[
		\P\left[\left\langle \x_{i},\frac{1}{\left|\calM\right|}\sum_{j\in\calM}\x_{j}\right\rangle \ge\frac{3\rho^{2}\left(t\sqrt{\left|\calM\right|}+\vecdim+\log\left(1/\delta\right)\right)}{\left|\calM\right|}\right]\le\exp\left(-\min\left\{ \frac{t^{2}}{4\vecdim},\frac{t}{4}\right\} \right)+\delta.
		\]
	\end{lem}
	Taking $t=4C\left(\sqrt{\vecdim\log\num}+\log\num\right)$ and $\delta=\num^{-20}$
	(where $C>0$ is a sufficiently large universal constant), we prove
	the bound in Lemma~\ref{lem:prod_of_noise_with_sum_noise}.
\end{proof}
%\Halmos
%\endproof

\subsection{Random Vectors on the Unit Ball\label{sec:proof_ball_model_sgnorm}}

In this section we prove Fact~\ref{fact:ball_model_sgnorm} concerning
the sub-Gaussian norm of a rotationally invariant random vector on
the unit $\ell_{2}$ ball.
%\proof{\textit{Proof of Fact~\ref{fact:ball_model_sgnorm}. }}
\begin{proof}[Proof of Fact~\ref{fact:ball_model_sgnorm}]
	Let $\g=(g_{1},\ldots,g_{\vecdim})^{\top}$ be a random vector drawn
	from a rotationally invariant distribution supported on the unit $\ell_{2}$
	ball in $\real^{\vecdim}$. Also let $C,C'>0$ be universal constants
	whose values may change line by line. By \cite[Proposition 4.10]{bilodeau2008mutivariate},
	$\g$ can be represented as $\g\overset{\textup{d}}{=}r\u,$ where
	$\overset{\textup{d}}{=}$ means equality in distribution, $r\overset{\textup{d}}{=}\norm[\g]2$,
	$\u$ is uniformly distributed on the unit sphere in $\real^{\vecdim}$,
	and $r$ and $\u$ are independent. The random vector $\u$ is sub-Gaussian
	with norm $\norm[\u]{\psi_{2}}\le C\sqrt{\frac{1}{\vecdim}}$ \cite[Theorem 3.4.5]{vershynin2017high},
	so its one-dimensional margin satisfies $\P\left\{ \left|u_{1}\right|>t\right\} \le2\exp\left(-\frac{t^{2}}{C'/\vecdim}\right).$
	On the other hand, we have $r\overset{\textup{d}}{=}\norm[\g]2\in[0,1]$
	since $\g$ is supported on the unit ball. Putting together the above
	facts gives 
	\[
	\P\left\{ \left|g_{1}\right|>t\right\} =\P\left\{ r\left|u_{1}\right|>t\right\} \le\P\left\{ \left|u_{1}\right|>t\right\} \le2\exp\left(-\frac{t^{2}}{C'/\vecdim}\right),
	\]
	whence $\norm[g_{1}]{\psi_{2}}\le C\sqrt{\frac{1}{\vecdim}}.$ By
	rotation invariance, we know that $\left\langle \a,\g\right\rangle \overset{\textup{d}}{=}g_{1}$
	for all unit vector $\a$ \cite[Proposition 4.8]{bilodeau2008mutivariate}.
	Therefore, all one-dimensional margins $\left\langle \a,\g\right\rangle $
	of $\g$ is sub-Gaussian with norm at most $C\sqrt{\frac{1}{\vecdim}}$.
	This completes the proof.
\end{proof}
%\Halmos
%\endproof
%\end{APPENDICES}

% Acknowledgments here
\section*{Acknowledgments}
% Enter the text of acknowledgments here
Y.\ Fei and Y.\ Chen are partially supported by NSF grant CCF-1704828 and CAREER Award CCF-2047910.
 
\bibliographystyle{plainnat}
\bibliography{references}

\end{document}